\documentclass[11pt]{article}

\usepackage[final]{acl}

\usepackage{times}
\usepackage{latexsym}
\usepackage{placeins}

\usepackage[T1]{fontenc}

\usepackage[utf8]{inputenc}

\usepackage{microtype}

\usepackage{inconsolata}

\usepackage{graphicx}

\usepackage{multirow}
\usepackage[normalem]{ulem}
\useunder{\uline}{\ul}{}
\usepackage{ulem}
\usepackage{todonotes}
\usepackage{comment}

\usepackage{enumitem}
\usepackage{amsmath}
\usepackage{array}  
\usepackage[normalem]{ulem}

\usepackage{booktabs}
\usepackage{CJKutf8}
\usepackage[T1]{fontenc}
\usepackage[utf8]{inputenc}

\usepackage{graphicx}   %
\usepackage{multirow}   %

\usepackage[table]{xcolor}

\usepackage{tikz}
\usetikzlibrary{arrows.meta, positioning, shapes.geometric, calc}

\title{SteerEval: Inference-time Interventions \\Strengthen Multilingual Generalization in Neural Summarization Metrics}

\author{
  \textbf{Silvia Casola\textsuperscript{1}}\thanks{~~Equal contribution. Ordering decided on a coin flip. \\ The work of Silvia Casola was carried out while at MaiNLP, LMU Munich \& Munich Center for Machine Learning.},
  \textbf{Ryan Soh-Eun Shim\textsuperscript{2}}\footnotemark[1],
  \\
  \textbf{Felicia Körner\textsuperscript{2}},
  \textbf{Yuchen Mao\textsuperscript{3}},
  \textbf{Barbara Plank\textsuperscript{2}}
\\
\\
  \textsuperscript{1}LaSTUS Lab, TALN Group, Universitat Pompeu Fabra (UPF), Barcelona, Spain \\
  \textsuperscript{2}MaiNLP, Center for Information and Language Processing, LMU Munich, Germany \\ \& Munich Center for Machine Learning \\
  \textsuperscript{3}Language Science and Technology, Saarland University, Germany
\\
}

\begin{document}
\maketitle

\begin{abstract}
An empirical bottleneck for Natural Language Generation tasks such as summarization is the lack of accurate and robust \emph{evaluation metrics for many languages}, which hinders progress. While Large Language Models (LLMs) are increasingly adopted as evaluators with the advent of LLM-as-a-Judge approaches, their agreement with human judgments remains low across many languages. We take inspiration from recent studies suggesting that multilingual language models often use English as an internal pivot language, and that misalignment with this pivot can lead to degraded downstream performance during text generation. Motivated by this hypothesis, we ask whether steering the activations of neural metrics with respect to an English pivot can improve their correlation with human judgments in multilingual summarization evaluation. We experiment with encoder- and decoder-based metrics across five languages. We find that activation steering is effective for neural metrics, and that a single, language-independent steering vector improves metric–human agreement. Steering also results in improvements when applied to COMET, a small encoder-based metric, showing notable human agreement despite its small scale.

\end{abstract}

\section{Introduction}
Metrics for Natural Language Generation (NLG) tasks such as summarization have become increasingly model-based \cite{Zhang*2020BERTScore, rei-etal-2020-comet, sellam-etal-2020-bleurt}, with the latest trend leveraging Large Language Models (LLMs) as evaluators \cite{liu-etal-2023-g, fu-etal-2024-gptscore}, a paradigm know as LLM-as-a-Judge \cite{10.5555/3666122.3668142}. 

When built on multilingual backbones, LLM evaluators can generalize across languages to some extent \cite{Zhang*2020BERTScore, rei-etal-2020-comet, sellam-etal-2020-bleurt}. However, preliminary studies on model-based multilingual evaluation suggest that they often do not correlate well with human judgments \cite{mondshine-etal-2025-beyond-n}, limiting their reliability as proxies for human evaluation.

We hypothesize that this gap is partly due to the way multilingual models process different languages: recent work has shown that multilingual LLMs internally rely on English as a pivot language \citep{wendler-etal-2024-llamas}, and that the degree of alignment between the internal representations of a language and this pivot varies between languages \cite{wang-etal-2025-bridging}. This misalignment has been linked to poor performance in multilingual factual recall \cite{lu-etal-2025-paths} and other discriminative and generative tasks \cite{wang-etal-2025-bridging, bafna2025translationbarrierhypothesismultilingual}, suggesting that explicitly controlling this alignment could improve performance.  \citet{wang-etal-2025-bridging}, for example, directly intervene on internal representations by steering them toward (or away from) English, showing this intervention is beneficial across a wide variety of NLG tasks. \citet{lu-etal-2025-paths} further show how alignment to the English pivot improves factual recall across languages. 

\begin{figure*}[t]
  \centering
    \includegraphics[width=\textwidth, ,
        trim=0 5 0 0,
        clip]{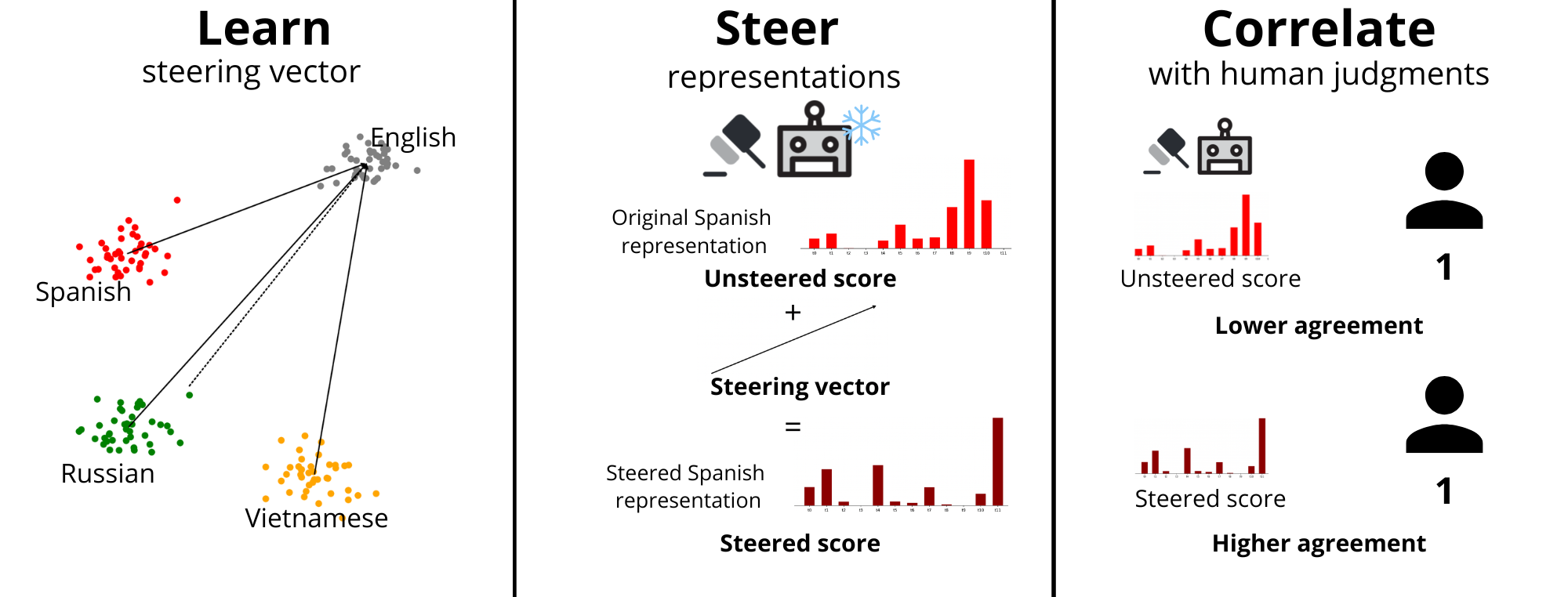}
  \caption{Illustration of our \textsc{SteerEval} method. We first use parallel data to learn cross-lingual vectors of mappings from \textit{Source language X $\rightarrow$ English} or a single multilingual average vector. We then apply the steering to neural metrics. Finally, we show that such steering results in better correlation with human judgments.}
  \label{fig:script-vector-pipeline}
\end{figure*}

Such work, however, is typically limited to discriminative and generative tasks: whether activation steering can improve performance on \textit{evaluation} tasks, and specifically on model-based metrics such as LLM-as-a-Judge, remains unexplored. This is a particularly important gap: reliable automatic metrics for multilingual evaluation are scarce, making the quality of model-based evaluation a bottleneck for progress across languages. As such, in this work we explicitly ask: To what degree does steering the activations of low-resource languages with respect to English strengthen \textit{LLM-as-a-Judge} performance?

To answer this question, we focus on summarization evaluation and first conduct a meta-analysis of LLM-as-a-Judge summarization metrics, assessing their reliability and alignment with human judgments. Corroborating prior work, we find modest and inconsistent correlation with human judgments \citep{mondshine-etal-2025-beyond-n}, showing the need for strengthened multilingual evaluators.
To this end, we investigate the effect of steering multilingual LLM-as-a-Judge models with respect to a high-resource pivot language at test time, and find it improves metric reliability. We further demonstrate that language-specific steering vectors may not be required, and a single, multilingual vector can lead to the largest improvements. Finally, we show that small encoder-based metrics such as COMET \citep{rei-etal-2020-comet} show surprisingly competitive performance, and that this performance can further be improved through steering. Our contributions are as follows:\footnote{Our code is available at: \url{https://github.com/mainlp/steereval}}

\paragraph{Multilingual metric meta-analysis.} We perform a meta-analysis of model-based summarization metrics across five typologically diverse languages. We examine two LLM-as-a-judge methods, using four backbone models and two prompt variations. We also examine COMET, an encoder-based metric, which shows competitive performance for summarization.

\paragraph{Activation steering for multilingual metrics.} We apply activation steering to two popular LLM-as-a-judge methods---based on next direct prompting \citep{liu-etal-2023-g} and average log likelihood \citep{fu-etal-2024-gptscore}---and we show activation steering strengthens correlation with human judgments for both approaches. We also extend steering to encoder-based models, largely unexplored in the broader activation steering literature, and we also observe consistent correlation gains. Our intervention consists of a simple and efficient vector-based addition, occurs at inference time, and requires no model retraining, making it an efficient way to strengthen metrics.

\paragraph{Multilingual intervention} We measure the cross-lingual similarity between \textit{Source language X} $\rightarrow$ \textit{English} intervention vectors in LLMs, and find these vectors to be largely similar across languages, corroborating prior work in the generalizability of refusal vectors \citep{wang2025refusaldirectionuniversalsafetyaligned} and, more broadly, shared language geometries \cite{chi-etal-2020-finding, dumas-etal-2025-separating}. Building on this finding, we explore multilingual steering, where we average over language-specific vectors to obtain a single multilingual steering vector. We show this solution outperforms language-specific steering across many settings.

\section{Related Work}
\subsection{Multilingual summarization evaluation}
Automatic evaluation of natural language generation (NLG) and summarization is a longstanding challenge \cite{10.1145/3485766, celikyilmaz2021evaluationtextgenerationsurvey}. Most existing work has focused on English. For instance, \citet{fabbri-etal-2021-summeval} evaluate 14 summarization metrics across four quality dimensions, while subsequent studies show that correlation with human judgments of reference-based metrics can drop when scoring LLM-generated outputs given their high quality \cite{casola-etal-2025-references}. As evaluation has increasingly shifted towards the LLM-as-a-Judge paradigm, \citet{bavaresco-etal-2025-llms} conducted a large-scale evaluation across multiple tasks, including summarization, and demonstrated high variability in metric behavior across different LLM judges.

Beyond English, automatic evaluation is understudied, and metric reliability often tends to further degrade. \citet{clark-etal-2023-seahorse} introduce a multilingual dataset in which model outputs are annotated in a binary fashion for specific desirable properties (e.g., conciseness), and shows it can be used to train ad-hoc evaluation metrics. Similarly, \citet{hada-etal-2024-metal} propose a multilingual dataset with human judgments using either binary labels or a 0--2 scale, and use it to assess LLMs as evaluators across five dimensions. More focused studies include \citet{barnes2025summarizationmetricsspanishbasque}, who analyze LLM-as-a-judge performance for Spanish and Basque, highlighting variation in evaluators' behaviors. Finally, \citet{mondshine-etal-2025-beyond-n} recently introduced an abstractive summarization dataset for meta-evaluating automatic metrics across eight diverse languages, focusing on n-gram- and small model-based metrics, concluding that correlations with human judgments remain relatively low in multilingual settings.

\subsection{Activation steering}
\label{sec:previous_steering}
The linear representation hypothesis \cite{10.5555/3692070.3693675} proposes that high-level concepts are represented as directions in the activation space of language models. Motivated by this hypothesis, an increasing body of work has used a test-time method known as activation steering to control LLM behavior. Such works typically isolate the direction that represents a given concept by vector arithmetic, and the isolated vector is then added to activations at test time \citep{turner2024steeringlanguagemodelsactivation}. Such a steering method has been applied to induce or reduce characteristics such as truthfulness \citep{marks2024geometrytruthemergentlinear}, toxicity, sentiment \citep{turner2024steeringlanguagemodelsactivation}, refusal \citep{arditi2024refusallanguagemodelsmediated}, and language steering \citep{sterz-etal-2025-recover}. 

Beyond steering vectors, \citet{wang-etal-2025-bridging} show that a learned linear map between low- and high-resource language representations can serve as an effective steering mechanism. More broadly, intervention methods can use other techniques, such as representation editing \cite{NEURIPS2022_6f1d43d5} and prompt optimization \cite{pryzant-etal-2023-automatic}.

In this paper, we focus on steering, and specifically on the vector-based approach, which we apply to strengthening multilingual summarization evaluation on non-English languages.

\section{Experimental Setup}

\subsection{Vector-based metric steering}
\label{sec:steering}
Figure \ref{fig:script-vector-pipeline} provides an overview of our steering methodology.
Specifically, we perform a vector-based intervention via the following steps.

\paragraph{Steering direction learning} 

We examine two alternative approaches: a monolingual intervention for a specific source language (Section \ref{sec:results}) and a multilingual intervention for all languages (Section \ref{sec:analysis_cross}).
For the monolingual intervention, we derive a steering direction between a source language and English from parallel data. Specifically, given representations in the source (lower-resourced) language $h_s$ and English (the target language) $h_t$, we compute a per-layer steering vector by subtracting the mean of $h_s$ from the mean of $h_t$ over all time steps, capturing the direction from the source to the target language in activation space:
\begin{equation*}
    \bar{h}_{v,l} = \bar{h}_{t,l} - \bar{h}_{s,l}
\end{equation*}
We also experiment with a single, multilingual vector, mapping from all languages to English, in line with previous work \cite{lu-etal-2025-paths}. To do so, we average across language-specific vectors, and obtain our final steering vector as: 

$$\bar{h}_{v,l} = \frac{1}{|M|} \sum_{m \in M} \left( \bar{h}_{t,l} - \bar{h}_{m,l} \right)$$
where $m$ is a source language in our language set $M$ and $|M|=5$ in our experiments. 

All representations are taken from the residual stream at the output of layer $l$, i.e., the hidden state produced by block $l$ before it is passed to block $l+1$.
Note that the parallel data used to learn the steering vector is disjoint from the summarization evaluation data, and any parallel corpus can in principle be used for this purpose.

\paragraph{Test time intervention} Keeping the language model frozen, we then steer its internal representations.
Specifically, at test time, we add the monolingual or multilingual steering vector to the activations of a single layer $l$:
\begin{equation*}
    h^{\text{steered}}_{q,l} = h_{q,l} + \sigma \, \bar{h}_{v,l}
\end{equation*}
where $h_{q,l}$ is the unsteered representation of test input $q$ at layer $l$, and $\sigma$ is a scalar parameter controlling the strength and sign of the steering.

For language-specific steering (Section \ref{sec:results}), for every target language, we jointly tune the steering strength $\sigma$ and the application layer $l$ on a language-specific development set with human judgments, and then evaluate the selected configuration on the corresponding disjoint test set. Following prior work \cite{lu-etal-2025-paths, bafna2025translationbarrierhypothesismultilingual}, we restrict the layer search space to the middle and final layers, as these exhibit the greatest representational mismatch with English for other tasks. For COMET, we steer the pooling-space embedding directly (no per-layer selection). We search over $\sigma \in [-5, 5]$ with a step size of 1. Hyperparameters are selected via joint grid search to maximize the average Pearson correlation across six evaluation dimensions (see Section \ref{sec:data}). 
In the multilingual vector experiments (Section \ref{sec:analysis_cross}), we apply the steering vector zero-shot without hyperparameter tuning.

\paragraph{Performance testing} We evaluate the metric using the steered representations and analyze its agreement with human judgments. Since the Seahorse dataset (\citealp{clark-etal-2023-seahorse}, which we discuss in Section \ref{sec:data}) is annotated with binary labels, we use Pearson correlation and ROC-AUC, following the original paper. 

\subsection{Datasets}
\label{sec:data}
\paragraph{Steering direction learning} To learn the language alignments, we leverage the FLORES multiparallel dataset \cite{flores}, which consists of human-translated sentences primarily from Wikipedia. We follow \citet{wang-etal-2025-bridging} and use 500 parallel sentence pairs.

\paragraph{Parameter tuning and performance testing} 

We use the Seahorse multilingual summarization dataset \cite{clark-etal-2023-seahorse} as key benchmark, a high-quality human evaluation dataset containing binary scores for six languages i.e., English, German, Spanish, Russian, Turkish, and Vietnamese. Source texts are extracted from four datasets in the GEM benchmark \cite{gehrmann-etal-2021-gem}: XSum (\citealp{narayan-etal-2018-dont}, for English only), XL-Sum \cite{hasan-etal-2021-xl}, MLSum \cite{scialom-etal-2020-mlsum}, and WikiLingua \cite{ladhak-etal-2020-wikilingua}. Summaries are the outputs of eight abstractive summarization systems based on T5 (\citealp{raffel2023exploringlimitstransferlearning}, only for English), mT5 \cite{xue-etal-2021-mt5} and PaLM \cite{chowdhery2022palmscalinglanguagemodeling}; human references from the respective datasets, when available, are also treated as hypotheses. The dataset is large-scale, with language-specific test sets of 2,600+ annotated outputs. Each summary is annotated by trained raters on six binary dimensions: whether it is a) comprehensible, b) non-repetitive, c) grammatical, d) concise, e) whether its content is fully attributable to the source article, and f) whether it captures the main
ideas of the source article. We preserve the original split and use the dataset validation set to tune the method hyperparameters and its test set for the final evaluation.\footnote{We provide dataset statistics in Appendix \ref{app:data}.}

\subsection{Metrics}
\label{sec:metrics}
We primarily examine the following metrics:
\paragraph{Direct prompting} We directly prompt a multilingual model to produce a score for each evaluation dimension. However, directly extracting the score from the model output can be unreliable, as responses may include additional text or justifications, an issue even more complex in multilingual settings. Thus, we rely on token probabilities. We follow G-Eval \cite{liu-etal-2023-g} and compute the score as a weighted sum over the token probabilities corresponding to the evaluation scale at the first generated token. The score is computed as: 
$$\text{score} = \sum_i p(s_i) \times s_i $$
where \( s_i \in \{0, 1\}  \), mimicking the dataset annotation, and $p(s_i)$ is the token probability. 

\paragraph{GPTScore (prompt-conditioned log-likelihood)}
Following ~\citet{fu-etal-2024-gptscore}, we evaluate a hypothesis using the model's conditional generation probabilities. Given a prompt that specifies a desired property (e.g., conciseness), the model is instructed to generate a summary exhibiting that property. The summary to be evaluated is thus scored by computing the average log-probability of its tokens under the model, conditioned on the prompt. Specifically, we compute the final score as: 
$$\text{score} =
\frac{1}{N} \sum_{t=1}^{N}  \log p\!\left(
h_t \mid h_{<t}, P, \theta
\right) $$ 
where $P$ is the prompt, $N$ is the number of tokens in the summary, and $\theta$ the model parameters. 

For both LLM-based metrics, we use Llama-3.2-3B\footnote{\textit{meta-llama/Llama-3.2-3B}} \cite{grattafiori2024llama3herdmodels} and aya-expanse-8b\footnote{\textit{CohereLabs/aya-expanse-8b}} \cite{dang2024ayaexpansecombiningresearch}, as backbones and further  experiment with their larger counterparts i.e., Llama-3.1-8B Instruct\footnote{\textit{meta-llama/Llama-3.1-8B-Instruct}} and aya-expanse-32b\footnote{\textit{CohereLabs/aya-expanse-32b}} in Section \ref{sec:analysis_cross}. These models are chosen since they are open, instruction-tuned, diverse in size, have broad language coverage, and have shown good performance in various multilingual tasks. We prompt each model both in English and with a machine translated version in the target language. Appendix \ref{tab:gptscore_prompts} details the metrics' implementation, including the prompts. 

\paragraph{COMET}
Due to prior work reporting state-of-the-art performance on many languages by directly applying COMET on multilingual summarization evaluation \citep{mondshine-etal-2025-beyond-n, krubinski-pecina-2022-comet}, we also analyze the effect of steering on COMET, which is a multilingual neural metric typically used in machine translation. We rely on the \textit{wmt22-comet-da} model, which is built on top of XLM-R \cite{conneau-etal-2020-unsupervised} and follow the experimental design of \citet{mondshine-etal-2025-beyond-n} to adapt the machine-translation metric to summarization: we replace the source input with an empty string, use the system-generated summary as the hypothesis, and the human-written summary as the reference. 
Note that COMET, in contrast to LLM-based judges, requires a reference summary. In the Seahorse dataset, reference summaries are themselves treated as hypotheses and scored by human judges. We therefore compute COMET scores only for the subset of model-generated hypotheses that have an associated reference, and results are not directly comparable to those from the LLMs.

\section{Language-specific steering}
\label{sec:results}
\begin{table*}[t]
\centering
\resizebox{\textwidth}{!}{%
\begin{tabular}{ll|c|c|c|c|c}
\textbf{} & \textbf{} & \textbf{de} & \textbf{es} & \textbf{ru} & \textbf{tr} & \textbf{vi} \\
\toprule
\multicolumn{7}{l}{\textbf{GPTScore}} \\
\midrule
\multirow{2}{*}{\texttt{Llama-3.2-3B}} & Baseline (EN-prompt) & $.022_{\pm.132}$& $.213_{\pm.141}$ & $.270_{\pm.062}$ & $.202_{\pm.108}$ & $\underline{.139}_{\pm.092}$ \\
 & + vector steer (EN-prompt) & {$.022_{\pm.127}$}& \textcolor{blue!70!black}{$.215_{\pm.142}$$\uparrow$}& \textcolor{blue!70!black}{$.276_{\pm.062}$$\uparrow$}& \textcolor{blue!70!black}{$\underline{.211}_{\pm.100}$$\uparrow$} & \textcolor{red!70!black}{$.138_{\pm.095}$$\downarrow$} \\
 & + avgvec zero-shot (EN-prompt) & \textcolor{blue!70!black}{$\underline{.024}_{\pm.127}$$\uparrow$} & \textcolor{blue!70!black}{$\underline{.221}_{\pm.149}$$\uparrow$} & \textcolor{blue!70!black}{$\underline{.281}_{\pm.062}$$\uparrow$}& \textcolor{blue!70!black}{$.210_{\pm.104}$$\uparrow$}& {$\underline{.139}_{\pm.097}$}\\
\midrule
\multirow{2}{*}{\texttt{aya-expanse-8b}} & Baseline (EN-prompt) & $.017_{\pm.143}$ & $.150_{\pm.088}$ & $.240_{\pm.067}$ & $.190_{\pm.069}$ & $.181_{\pm.063}$ \\
 & + vector steer (EN-prompt) & \textcolor{blue!70!black}{$.018_{\pm.145}$$\uparrow$}& \textcolor{blue!70!black}{$.155_{\pm.088}$$\uparrow$}& \textcolor{blue!70!black}{$.247_{\pm.067}$$\uparrow$}& \textcolor{blue!70!black}{$.198_{\pm.070}$$\uparrow$}& \textcolor{blue!70!black}{$.185_{\pm.063}$$\uparrow$}\\
 & + avgvec zero-shot (EN-prompt) & \textcolor{blue!70!black}{$\underline{.021}_{\pm.143}$$\uparrow$} & \textcolor{blue!70!black}{$\underline{.156}_{\pm.089}$$\uparrow$} & \textcolor{blue!70!black}{$\underline{.248}_{\pm.067}$$\uparrow$} & \textcolor{blue!70!black}{$\underline{.199}_{\pm.070}$$\uparrow$} & \textcolor{blue!70!black}{$\underline{.188}_{\pm.063}$$\uparrow$} \\
\midrule
\multirow{2}{*}{\texttt{Llama-3.1-8B-Instruct}} & Baseline (EN-prompt) & $.036_{\pm.138}$ & $.228_{\pm.133}$ & $.282_{\pm.059}$ & $.242_{\pm.100}$ & $\underline{.176}_{\pm.081}$ \\
 & + avgvec zero-shot (EN-prompt) & \textcolor{blue!70!black}{$\underline{.040}_{\pm.133}$$\uparrow$}& \textcolor{blue!70!black}{$\mathbf{\underline{.238}}_{\pm.143}$$\uparrow$} & \textcolor{blue!70!black}{$\mathbf{\underline{.294}}_{\pm.058}$$\uparrow$} & \textcolor{blue!70!black}{$\mathbf{\underline{.250}}_{\pm.096}$$\uparrow$} & \textcolor{red!70!black}{$.175_{\pm.085}$$\downarrow$} \\
\midrule
\multirow{2}{*}{\texttt{aya-expanse-32b}} & Baseline (EN-prompt) & $.007_{\pm.139}$ & $.164_{\pm.084}$ & $.244_{\pm.067}$ & $.190_{\pm.070}$ & $.195_{\pm.053}$ \\
 & + avgvec zero-shot (EN-prompt) & \textcolor{blue!70!black}{$\underline{.011}_{\pm.140}$$\uparrow$} & \textcolor{blue!70!black}{$\underline{.169}_{\pm.085}$$\uparrow$} & \textcolor{blue!70!black}{$\underline{.250}_{\pm.065}$$\uparrow$} & \textcolor{blue!70!black}{$\underline{.196}_{\pm.070}$$\uparrow$} & \textcolor{blue!70!black}{$\underline{.199}_{\pm.053}$$\uparrow$}\\
\midrule
\multicolumn{7}{l}{\textbf{Direct Prompting}} \\
\midrule
\multirow{2}{*}{\texttt{Llama-3.2-3B}} & Baseline (EN-prompt) & $\underline{.050}_{\pm.050}$ & $-.004_{\pm.045}$ & $.014_{\pm.039}$ & $.007_{\pm.043}$ & $-.001_{\pm.040}$ \\
 & + vector steer (EN-prompt) & \textcolor{red!70!black}{$.049_{\pm.048}$$\downarrow$} & \textcolor{blue!70!black}{$\underline{-.001}_{\pm.044}$$\uparrow$} & \textcolor{blue!70!black}{$\underline{.015}_{\pm.039}$$\uparrow$} & \textcolor{blue!70!black}{$\underline{.008}_{\pm.045}$$\uparrow$} & \textcolor{blue!70!black}{$\underline{.005}_{\pm.040}$$\uparrow$} \\
\midrule
\multirow{2}{*}{\texttt{aya-expanse-8b}} & Baseline (EN-prompt) & $\mathbf{\underline{.208}}_{\pm.042}$ & $\underline{.226}_{\pm.139}$& $.249_{\pm.079}$ & $.237_{\pm.112}$ & $\mathbf{\underline{.248}}_{\pm.077}$ \\
 & + vector steer (EN-prompt) & {$\mathbf{\underline{.208}}_{\pm.042}$} & {$\underline{.226}_{\pm.139}$}& \textcolor{blue!70!black}{$\underline{.251}_{\pm.076}$$\uparrow$} & \textcolor{blue!70!black}{$\underline{.239}‚_{\pm.113}$$\uparrow$}& \textcolor{red!70!black}{$.247_{\pm.077}$$\downarrow$} \\
\bottomrule
\end{tabular}%
}
\caption{SEAHORSE Pearson $r$ averaged over 6 quality dimensions (Comprehensible, Non-repetitive, Grammatical, Attributable, Main ideas, Concise) with standard deviation as subscripts ($\text{mean}_{\pm\text{std}}$), per language, using English prompts. Baseline uses $\sigma{=}0$ (no steering); steered uses the best $\sigma$ selected on the validation split. \textbf{Bold} indicates the single highest value per language across the entire table; \underline{underline} indicates the highest value per language within each model. \textcolor{blue!70!black}{Blue}/$\uparrow$ (\textcolor{red!70!black}{red}/$\downarrow$) on steered rows indicates improvement (resp.\ degradation) relative to the baseline.}
\label{tab:seahorse_condensed_en}
\end{table*}

Tables \ref{tab:seahorse_condensed_en} and \ref{tab:seahorse_comet_condensed} summarize the main results of our LLM-as-a-Judge and COMET steering experiments respectively. In this section, we focus on language-specific steering and later discuss the zero-shot multilingual average vector setup in Section \ref{sec:analysis_cross}. We primarily report results using Pearson correlation and focus on English prompting in the main text; detailed results when prompting in the source language are in Appendix \ref{app:src_results}, while results using AUC are presented in Appendix \ref{app:auc_results}. 
We report the average correlation over the six evaluation dimensions per language; per-dimension results are in Appendix \ref{app:dimensions}.

\begin{table*}[t]
  \centering
  \resizebox{0.7\textwidth}{!}{
  \begin{tabular}{l|c|c|c|c|c}
  & \textbf{de} & \textbf{es} & \textbf{ru} & \textbf{tr} & \textbf{vi} \\
  \toprule
  \multicolumn{6}{l}{\textbf{COMET}} \\
  \midrule
  Baseline & $.248_{\pm.087}$ & $.266_{\pm.120}$ & $.358_{\pm.130}$ & $.380_{\pm.139}$ & $.290_{\pm.116}$ \\
  + vector steer & \textcolor{blue!70!black}{$\mathbf{.279}_{\pm.081}$$\uparrow$} &
  \textcolor{blue!70!black}{$\mathbf{.318}_{\pm.184}$$\uparrow$} & \textcolor{red!70!black}{$.351_{\pm.155}$$\downarrow$} &
  \textcolor{red!70!black}{$.360_{\pm.144}$$\downarrow$} & \textcolor{blue!70!black}{$.295_{\pm.136}$$\uparrow$} \\
  + avgvec zero-shot & \textcolor{blue!70!black}{$.275_{\pm.084}$$\uparrow$} & \textcolor{blue!70!black}{$.303_{\pm.146}$$\uparrow$} &
  \textcolor{blue!70!black}{$\mathbf{.364}_{\pm.152}$$\uparrow$} & \textcolor{blue!70!black}{$\mathbf{.392}_{\pm.165}$$\uparrow$} &
  \textcolor{blue!70!black}{$\mathbf{.297}_{\pm.134}$$\uparrow$} \\
  \bottomrule
  \end{tabular}
  }
  \caption{SEAHORSE Pearson $r$ over  6 quality dimensions (Comprehensible, Non-repetitive, Grammatical, Attributable,
  Main ideas, Concise)  with standard deviation as subscripts ($\text{mean}_{\pm\text{std}}$), per language for COMET (\texttt{wmt22-comet-da}). Baseline uses $\sigma{=}0$ (no steering); steered uses the
  best $\sigma$ selected on the validation split. \textbf{Bold} indicates the highest value per language.
  \textcolor{blue!70!black}{Blue}/$\uparrow$ (\textcolor{red!70!black}{red}/$\downarrow$) indicates improvement (resp. degradation) subscripts
  relative to the baseline.}
  \label{tab:seahorse_comet_condensed}
  \end{table*}

\paragraph{Baselines show relatively modest and variable correlation with human judgments.} Across all settings, baseline systems achieve relatively low positive Pearson correlation, showing that summary evaluation remains challenging in multilingual settings, and that metrics designed for English do not necessarily transfer across languages, corroborating previous results \cite{clark-etal-2023-seahorse, mondshine-etal-2025-beyond-n}. Larger models do not consistently outperform smaller ones: for GPTScore, aya-expanse-32b shows comparable baseline performance to its smaller (8b) counterpart and, interestingly, lower performance than Llama-3.1-8b across all languages. No single configuration is superior across all settings: the choice of metric, backbone model, and prompt language interact in non-trivial ways. While comparisons with LLM-based scores should be taken with care given the modified test set (see Section \ref{sec:metrics}), COMET typically achieves a strong baseline correlations across all languages, including for languages like German that show consistently low correlations for LLM-based metrics. This is surprising given that COMET is a much smaller model trained specifically for translation quality estimation rather than summarization evaluation. We hypothesize this improvement might be due to the specific supervised training and to the use of a reference, not available to the LLMs in our prompts. 

\paragraph{Steering consistently improves performance.} Language-specific steering typically improves over the baseline, with gains in the vast majority of languages, metrics, and backbone models. While the advantage is generally incremental, the method shows consistent improvement across the majority of the configurations, which is notable given the simplicity and generality of our inference-time steering method. 
Steering typically improves over the baseline across LLM-based metrics, though the magnitude and consistency of gains vary considerably depending on the configuration. Improvements are also observed for COMET, though these are less consistent over languages when using language-specific steering.

\paragraph{GPTScore tends to be more stable across backbone models, and benefits most consistently from steering.} GPTScore tends to show more consistent results than Direct Prompting across backbone models. We hypothesize this stability might be due to the log-likelihood signal being more robust across model sizes when compared to prompting for direct scoring. Interestingly, the Llama-3B model shows typically even better GPTScore baseline performance than aya-8B across languages despite the difference in size,  suggesting that GPTScore may be less sensitive to model capacity than Direct Prompting. Steering improves GPTScore across nearly all language and model combinations: aya-expanse-8b and aya-expanse-32b show gains across all five languages with no degradations, while Llama-3.2-3B and Llama-3.1-8B-Instruct show only a single marginal degradation each, in Vietnamese. This suggests that the conditional log-likelihood scoring of multilingual models is particularly steerable, regardless of model size.

\paragraph{Direct prompting requires higher model capacity, and steering results are more mixed.} In contrast to GPTScore, direct prompting appears to require models with higher capacity to produce meaningful signal. For Llama-3.2-3B, direct prompting degrades to near-zero or even negative correlations when prompting in English, with baseline values of -.004 for Spanish and -.001 for Vietnamese, likely due to the backbone's limited ability to follow instructions. Despite this weak baseline, steering recovers some signal across nearly all languages. With a larger model, direct prompting is the best-performing configuration, with aya-expanse-8b reaching the highest baseline correlations across four out of five languages. However, steering results in limited improvement over baselines, producing marginal gains for Russian and Turkish, with a slight degradation for Vietnamese.

\paragraph{COMET shows strong but less consistent gains from language-specific steering.} As discussed above, COMET baselines are substantially higher than LLM-based metrics across all languages. Language-specific steering improves German and Spanish substantially, but degrades Russian and Turkish, with a marginal gain for Vietnamese. Note that the multiligual variant, which we discuss in Section \ref{sec:analysis_cross}, shows a more consistent picture, improving all five languages and achieving the best results for Russian, Turkish, and Vietnamese.

\paragraph{Performance across languages does not follow resourceness levels.} German, generally considered a relatively high-resource language, consistently shows the lowest correlations across all configurations. Under Direct Prompting with aya-expanse-8b, differences across languages are small, suggesting that larger and more multilingual models may be more robust to language variation. Turkish shows notable variability: it is among the strongest languages under Direct Prompting with aya-expanse-8b, but weaker under GPTScore with Llama-3.2-3B, possibly due to its morphological complexity (which  might be challenging for token log-likelihood scoring). When steering is applied, degradations are rare and typically marginal, concentrated in Vietnamese and German for LLMs. The pattern is similar for COMET, where the multilingual variant reverses the two degradations under vector steering (Russian and Turkish).

\paragraph{The effect of prompt language depends on the model and metric.} Prompting in the source language (Appendix \ref{app:src_results}) does not show a consistent advantage over English. Prompting in English and the source language results in different baseline strengths depending on the metrics: source languages typically produce stronger Direct Prompting baselines, while English tends to yield higher GPTScore baselines. Steering is more consistently beneficial when prompting in English.

\paragraph{Performance is highly variable across dimensions} Looking at Appendix \ref{app:dimensions}, baseline difficulty varies across dimensions. On average, Comprehensible is the easiest dimension, while Grammatical is the hardest one. One notable interaction concerns the metric: COMET substantially outperforms GPTScore and Direct Prompting on fluency-adjacent dimensions, but this advantage largely disappears for Grammatical  and reverses for Attributable, where GPTScore outperforms COMET. Language-specific vector steering typically leads to larger gains on Comprehensible and Non-repetitive.

\section{Multilingual steering}
\label{sec:analysis_cross}

Prior work in activation steering has found high level-concepts such as refusal or script to be broadly similar vectors across languages \citep{wang2025refusaldirectionuniversalsafetyaligned, shim2026linearscriptrepresentationsspeech, lu-etal-2025-paths}. In this section, we investigate to what extent language-specific steering is needed for model-based evaluation metrics. We focus on GPTScore and COMET, as they show more consistent and interpretable behaviors than direct prompting for monolingual steering. For simplicity, we experiment with prompting in English only.

\begin{figure*}[t]
    \centering
    \includegraphics[width=0.51\textwidth]{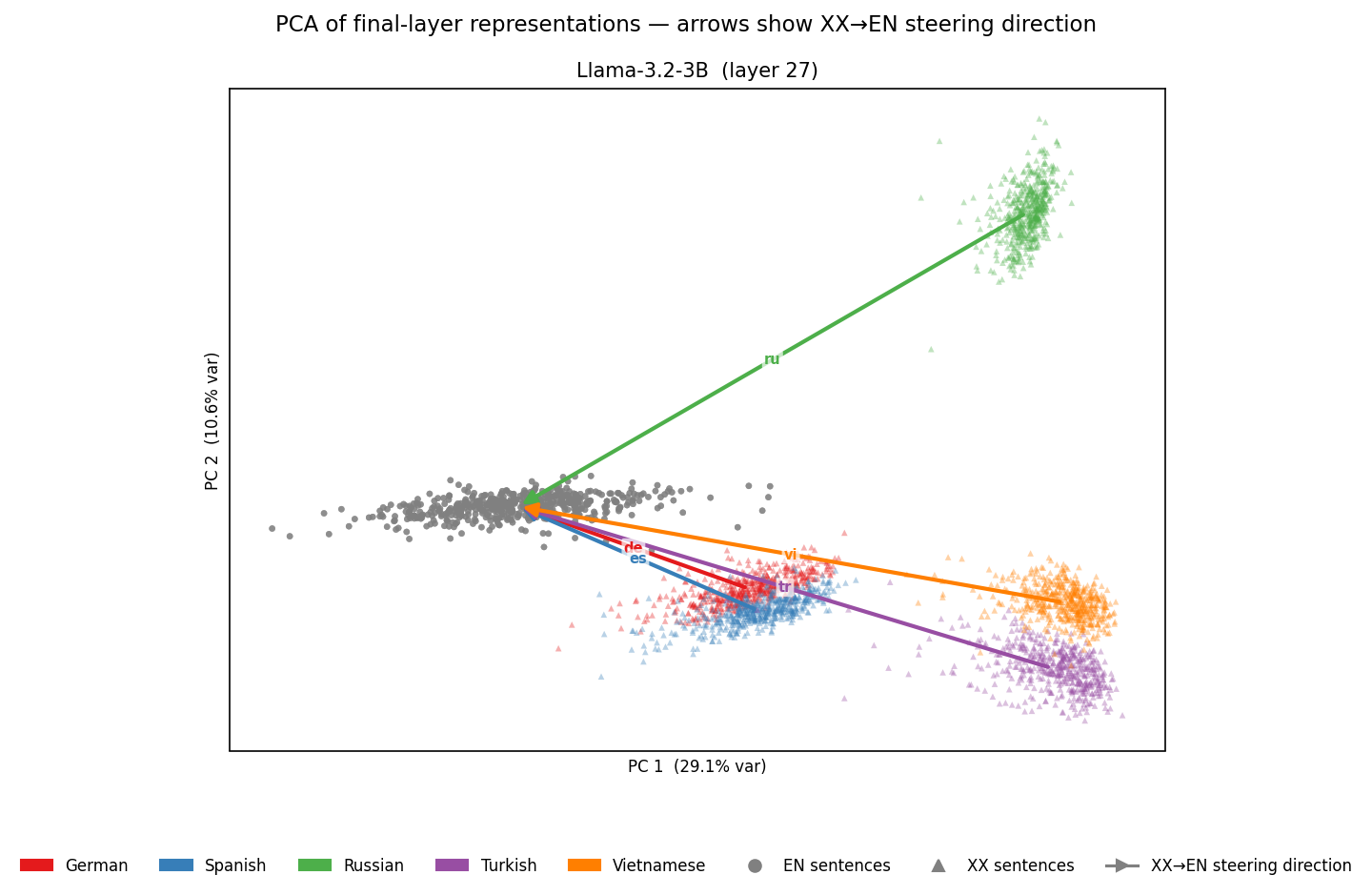}
    \includegraphics[width=0.4\textwidth]{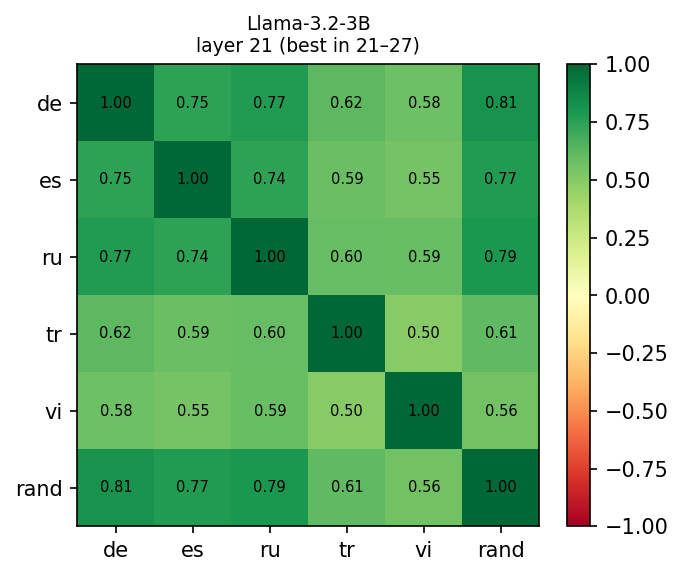}
    \caption{Projection of the instances from FLORES into the Principal Components (left) and cosine similarity across steering vectors (right) for Llama-3.2-3B.}
    \label{fig:pca_vec}
\end{figure*}

To study the similarity of cross-language steering vectors, we measure the extent to which \textit{Language X $\rightarrow$ English} directions are similar. As such, we compute the cosine similarity between our \textit{Language X $\rightarrow$ English} vectors extracted from parallel data.  
Figure~\ref{fig:pca_vec} shows the results of projecting language samples from the parallel data onto the principal components (PCA, left) and the cross-language similarity matrix (right) for GPTScore, for the Llama-3.2-3b model. We focus on the last model layer, as it is consistently identified as optimal on our development set across languages and settings. The PCA reveals that samples cluster by language, suggesting that the model's internal representations encode language identity at this layer. The similarity matrix, on the other hand, shows that the \textit{Language X $\rightarrow$
English} steering vectors are relatively similar across languages.

Motivated by this similarity, and in line with previous work \cite{lu-etal-2025-paths}, we explore the effect of taking a single steering vector across all languages, as detailed in Section \ref{sec:steering}. 

For GPTScore, we steer the last layer of the model, following previous observations;  for COMET, similarly to the language-specific steering, we directly steer the pooling-space embedding directly. We set $\sigma$ to -2.5 (the midpoint of our explored negative interval) for all languages, since the parameter is consistently negative across language.

Surprisingly, averaged zero-shot vector is at least as good as, and often better than, the language-specific tuned vector in nearly all configurations. Focusing on LLM-based metrics, Table \ref{tab:seahorse_condensed_en} (focusing on the avgvec zero-shot rows) shows the results. For Llama-3.2-3B, improvements over the baseline are observed for all languages except Vietnamese (for which results are consistent with no intervention), and the averaged vector results in higher correlations than the language-specific tuned vector for all languages except Turkish. For aya-expanse-8b, improvements are consistent across all languages, always exceeding language-specific performance. 

The effectiveness of a single multilingual vector is particularly interesting given that we performed no parameter tuning and pre-selected an average negative $\sigma$, a setting that is simpler than the language-specific one. Since we do not search for hyperparameters and use a single language-independent average vector, we can also scale to larger models with limited computational cost. Thus, we also explore whether our approach improves for larger models of the same family. After steering, GPTScore with Llama-3.1-8B achieves the best performance overall in Spanish, Russian, and Turkish, improving on non-trivial baselines; Vietnamese is the only language showing minor degradation, consistently with previous findings. 
aya-expanse-32b also shows improvement with respect to its baseline.

Improvements also stand for COMET, for which results are shown in Table \ref{tab:seahorse_comet_condensed}. When using a cross-lingual vector (without parameter tuning), COMET results are remarkably strong. Despite requiring no language-specific tuning, it matches or exceeds the language-specific steered variant in four out of five languages, and achieves the best overall result for Russian ($r=.364$), Turkish ($r=.392$), and Vietnamese ($r=.297$). Also notice that, unlike the language-specific vector, the cross-lingual vector never degrades performance: for Russian and Turkish, where individual steering vectors hurt, the cross-lingual vector not only recovers but exceeds the baseline, suggesting that averaging across languages might have a natural regularizing effect that suppresses language-specific noise.
Taken together, these results suggest that encoder-based metrics have strong potential when references are available, and can be steered effectively, making our approach generalizable beyond large decoder-based LLMs.

These results suggest that language-specific steering vectors might not be necessary: a single averaged vector, applied zero-shot with a fixed hyperparameter, might be sufficient to obtain consistent gains. %

\section{Conclusion}
In this paper, we leverage the insight that LLMs internally use English as a pivot language \citep{wendler-etal-2024-llamas} to improve the reliability of neural metrics for multilingual summarization evaluation. We showed that test-time activation steering consistently improves metrics' correlation with human judgments across a diverse set of languages, metrics, and backbone models, with no model retraining required. Our findings contribute to the growing body of work that leverages model-internal representations for practical downstream tasks \citep{marks2024geometrytruthemergentlinear}, and extend it to the underexplored setting of automatic evaluation. 

Notably, we find that language-specific steering is not necessary: a single averaged multilingual vector, applied zero-shot with a fixed hyperparameter, is sufficient to obtain consistent gains and often outperforms its language-specific counterpart. 

We also apply our method to COMET: despite being a much smaller encoder-based model, steered COMET benefits strongly from our intervention and surpasses LLM-based evaluators in human agreement across multiple languages.

Several directions remain open for future work. Our approach assumes the availability of  parallel data to derive steering vectors; moreover, future work should study how far the multilingual vector generalizes to unseen languages not included in the averaging set. Finally, while our work focuses on language steering, prior work has leveraged test-time interventions targeting specific evaluative dimensions such as fluency \citep{sheng-etal-2024-repeval}; combining language- and dimension-specific interventions could yield further gains in multilingual evaluation.

\section*{Limitations}

Due to the relative lack of large-scale summarization evaluation datasets with human annotations, our study is limited by the number of available languages. For the specific dataset we employ \citep{clark-etal-2023-seahorse}, the additional constraint that this dataset only has binary annotations is a further limitation. 

We optimize our steering hyperparameters on the average performance improvement across dimensions. Future work should investigate to what extent setting dimension-specific parameters could improve performance. 

While we demonstrate the effectiveness of a single multilingual steering vector across the languages in our study, we do not evaluate its generalization to unseen languages, i.e., languages not in the averaging set. Whether the vector transfers to typologically distant or lower-resourced languages remains an open question.

Steering representations toward an English pivot may introduce unwanted biases, e.g., cultural biases. We leave a systematic analysis of such biases to future work.

Finally, we limit our analysis to a vector-based steering intervention. Future work should investigate different steering methods \cite{wang-etal-2025-bridging} and other representation interventions, such as representation editing \cite{NEURIPS2022_6f1d43d5} or prompt optimization \cite{pryzant-etal-2023-automatic}.

\section*{Ethical Statement}
We acknowledge the usage of LLMs in the partial assistance of figure creation, code development, and in strengthening the coherence of parts of the manuscript.

\section*{Acknowledgments}
We would like to thank the members of the MaiNLP lab for their valuable feedback. We recognize the support for Ryan Soh-Eun Shim, Felicia Körner, and Barbara Plank through the ERC Consolidator Grant DIALECT 101043235. Silvia Casola was supported by the European Union’s Horizon Europe research and innovation programme under the Marie Skłodowska-Curie grant agreement No 101282681 (GenEval).

\bibliography{custom}

@inproceedings{pryzant-etal-2023-automatic,
    title = "Automatic Prompt Optimization with ``Gradient Descent'' and Beam Search",
    author = "Pryzant, Reid  and
      Iter, Dan  and
      Li, Jerry  and
      Lee, Yin  and
      Zhu, Chenguang  and
      Zeng, Michael",
    editor = "Bouamor, Houda  and
      Pino, Juan  and
      Bali, Kalika",
    booktitle = "Proceedings of the 2023 Conference on Empirical Methods in Natural Language Processing",
    month = dec,
    year = "2023",
    address = "Singapore",
    publisher = "Association for Computational Linguistics",
    url = "https://aclanthology.org/2023.emnlp-main.494/",
    doi = "10.18653/v1/2023.emnlp-main.494",
    pages = "7957--7968"
}

@inproceedings{NEURIPS2022_6f1d43d5,
 author = {Meng, Kevin and Bau, David and Andonian, Alex and Belinkov, Yonatan},
 booktitle = {Advances in Neural Information Processing Systems},
 doi = {10.52202/068431-1262},
 editor = {S. Koyejo and S. Mohamed and A. Agarwal and D. Belgrave and K. Cho and A. Oh},
 pages = {17359--17372},
 publisher = {Curran Associates, Inc.},
 title = {Locating and Editing Factual Associations in GPT},
 url = {https://proceedings.neurips.cc/paper_files/paper/2022/file/6f1d43d5a82a37e89b0665b33bf3a182-Paper-Conference.pdf},
 volume = {35},
 year = {2022}
}

@inproceedings{10.5555/3666122.3668142,
author = {Zheng, Lianmin and Chiang, Wei-Lin and Sheng, Ying and Zhuang, Siyuan and Wu, Zhanghao and Zhuang, Yonghao and Lin, Zi and Li, Zhuohan and Li, Dacheng and Xing, Eric P. and Zhang, Hao and Gonzalez, Joseph E. and Stoica, Ion},
title = {Judging LLM-as-a-judge with MT-bench and Chatbot Arena},
year = {2023},
publisher = {Curran Associates Inc.},
address = {Red Hook, NY, USA},
booktitle = {Proceedings of the 37th International Conference on Neural Information Processing Systems},
articleno = {2020},
numpages = {29},
location = {New Orleans, LA, USA},
series = {NIPS '23}
}

@misc{shim2026linearscriptrepresentationsspeech,
      title={Linear Script Representations in Speech Foundation Models Enable Zero-Shot Transliteration}, 
      author={Ryan Soh-Eun Shim and Kwanghee Choi and Kalvin Chang and Ming-Hao Hsu and Florian Eichin and Zhizheng Wu and Alane Suhr and Michael A. Hedderich and David Harwath and David R. Mortensen and Barbara Plank},
      year={2026},
      eprint={2601.02906},
      archivePrefix={arXiv},
      primaryClass={cs.CL},
      url={https://arxiv.org/abs/2601.02906}, 
}

@misc{raffel2023exploringlimitstransferlearning,
      title={Exploring the Limits of Transfer Learning with a Unified Text-to-Text Transformer}, 
      author={Colin Raffel and Noam Shazeer and Adam Roberts and Katherine Lee and Sharan Narang and Michael Matena and Yanqi Zhou and Wei Li and Peter J. Liu},
      year={2023},
      eprint={1910.10683},
      archivePrefix={arXiv},
      primaryClass={cs.LG},
      url={https://arxiv.org/abs/1910.10683}, 
}

@inproceedings{ladhak-etal-2020-wikilingua,
    title = "{W}iki{L}ingua: A New Benchmark Dataset for Cross-Lingual Abstractive Summarization",
    author = "Ladhak, Faisal  and
      Durmus, Esin  and
      Cardie, Claire  and
      McKeown, Kathleen",
    editor = "Cohn, Trevor  and
      He, Yulan  and
      Liu, Yang",
    booktitle = "Findings of the Association for Computational Linguistics: EMNLP 2020",
    month = nov,
    year = "2020",
    address = "Online",
    publisher = "Association for Computational Linguistics",
    url = "https://aclanthology.org/2020.findings-emnlp.360/",
    doi = "10.18653/v1/2020.findings-emnlp.360",
    pages = "4034--4048"
}

@inproceedings{scialom-etal-2020-mlsum,
    title = "{MLSUM}: The Multilingual Summarization Corpus",
    author = "Scialom, Thomas  and
      Dray, Paul-Alexis  and
      Lamprier, Sylvain  and
      Piwowarski, Benjamin  and
      Staiano, Jacopo",
    editor = "Webber, Bonnie  and
      Cohn, Trevor  and
      He, Yulan  and
      Liu, Yang",
    booktitle = "Proceedings of the 2020 Conference on Empirical Methods in Natural Language Processing (EMNLP)",
    month = nov,
    year = "2020",
    address = "Online",
    publisher = "Association for Computational Linguistics",
    url = "https://aclanthology.org/2020.emnlp-main.647/",
    doi = "10.18653/v1/2020.emnlp-main.647",
    pages = "8051--8067"
}

@inproceedings{narayan-etal-2018-dont,
    title = "Don{'}t Give Me the Details, Just the Summary! Topic-Aware Convolutional Neural Networks for Extreme Summarization",
    author = "Narayan, Shashi  and
      Cohen, Shay B.  and
      Lapata, Mirella",
    editor = "Riloff, Ellen  and
      Chiang, David  and
      Hockenmaier, Julia  and
      Tsujii, Jun{'}ichi",
    booktitle = "Proceedings of the 2018 Conference on Empirical Methods in Natural Language Processing",
    month = oct # "-" # nov,
    year = "2018",
    address = "Brussels, Belgium",
    publisher = "Association for Computational Linguistics",
    url = "https://aclanthology.org/D18-1206/",
    doi = "10.18653/v1/D18-1206",
    pages = "1797--1807"
}

@inproceedings{gehrmann-etal-2021-gem,
    title = "The {GEM} Benchmark: Natural Language Generation, its Evaluation and Metrics",
    author = "Gehrmann, Sebastian  and
      Adewumi, Tosin  and
      Aggarwal, Karmanya  and
      Ammanamanchi, Pawan Sasanka  and
      Aremu, Anuoluwapo  and
      Bosselut, Antoine  and
      Chandu, Khyathi Raghavi  and
      Clinciu, Miruna-Adriana  and
      Das, Dipanjan  and
      Dhole, Kaustubh  and
      Du, Wanyu  and
      Durmus, Esin  and
      Du{\v{s}}ek, Ond{\v{r}}ej  and
      Emezue, Chris Chinenye  and
      Gangal, Varun  and
      Garbacea, Cristina  and
      Hashimoto, Tatsunori  and
      Hou, Yufang  and
      Jernite, Yacine  and
      Jhamtani, Harsh  and
      Ji, Yangfeng  and
      Jolly, Shailza  and
      Kale, Mihir  and
      Kumar, Dhruv  and
      Ladhak, Faisal  and
      Madaan, Aman  and
      Maddela, Mounica  and
      Mahajan, Khyati  and
      Mahamood, Saad  and
      Majumder, Bodhisattwa Prasad  and
      Martins, Pedro Henrique  and
      McMillan-Major, Angelina  and
      Mille, Simon  and
      van Miltenburg, Emiel  and
      Nadeem, Moin  and
      Narayan, Shashi  and
      Nikolaev, Vitaly  and
      Niyongabo Rubungo, Andre  and
      Osei, Salomey  and
      Parikh, Ankur  and
      Perez-Beltrachini, Laura  and
      Rao, Niranjan Ramesh  and
      Raunak, Vikas  and
      Rodriguez, Juan Diego  and
      Santhanam, Sashank  and
      Sedoc, Jo{\~a}o  and
      Sellam, Thibault  and
      Shaikh, Samira  and
      Shimorina, Anastasia  and
      Sobrevilla Cabezudo, Marco Antonio  and
      Strobelt, Hendrik  and
      Subramani, Nishant  and
      Xu, Wei  and
      Yang, Diyi  and
      Yerukola, Akhila  and
      Zhou, Jiawei",
    editor = "Bosselut, Antoine  and
      Durmus, Esin  and
      Gangal, Varun Prashant  and
      Gehrmann, Sebastian  and
      Jernite, Yacine  and
      Perez-Beltrachini, Laura  and
      Shaikh, Samira  and
      Xu, Wei",
    booktitle = "Proceedings of the First Workshop on Natural Language Generation, Evaluation, and Metrics (GEM)",
    month = aug,
    year = "2021",
    address = "Online",
    publisher = "Association for Computational Linguistics",
    url = "https://aclanthology.org/2021.gem-1.10/",
    doi = "10.18653/v1/2021.gem-1.10",
    pages = "96--120"
}

@inproceedings{xue-etal-2021-mt5,
    title = "m{T}5: A Massively Multilingual Pre-trained Text-to-Text Transformer",
    author = "Xue, Linting  and
      Constant, Noah  and
      Roberts, Adam  and
      Kale, Mihir  and
      Al-Rfou, Rami  and
      Siddhant, Aditya  and
      Barua, Aditya  and
      Raffel, Colin",
    editor = "Toutanova, Kristina  and
      Rumshisky, Anna  and
      Zettlemoyer, Luke  and
      Hakkani-Tur, Dilek  and
      Beltagy, Iz  and
      Bethard, Steven  and
      Cotterell, Ryan  and
      Chakraborty, Tanmoy  and
      Zhou, Yichao",
    booktitle = "Proceedings of the 2021 Conference of the North American Chapter of the Association for Computational Linguistics: Human Language Technologies",
    month = jun,
    year = "2021",
    address = "Online",
    publisher = "Association for Computational Linguistics",
    url = "https://aclanthology.org/2021.naacl-main.41/",
    doi = "10.18653/v1/2021.naacl-main.41",
    pages = "483--498"
}

@misc{chowdhery2022palmscalinglanguagemodeling,
      title={PaLM: Scaling Language Modeling with Pathways}, 
      author={Aakanksha Chowdhery and Sharan Narang and Jacob Devlin and Maarten Bosma and Gaurav Mishra and Adam Roberts and Paul Barham and Hyung Won Chung and Charles Sutton and Sebastian Gehrmann and Parker Schuh and Kensen Shi and Sasha Tsvyashchenko and Joshua Maynez and Abhishek Rao and Parker Barnes and Yi Tay and Noam Shazeer and Vinodkumar Prabhakaran and Emily Reif and Nan Du and Ben Hutchinson and Reiner Pope and James Bradbury and Jacob Austin and Michael Isard and Guy Gur-Ari and Pengcheng Yin and Toju Duke and Anselm Levskaya and Sanjay Ghemawat and Sunipa Dev and Henryk Michalewski and Xavier Garcia and Vedant Misra and Kevin Robinson and Liam Fedus and Denny Zhou and Daphne Ippolito and David Luan and Hyeontaek Lim and Barret Zoph and Alexander Spiridonov and Ryan Sepassi and David Dohan and Shivani Agrawal and Mark Omernick and Andrew M. Dai and Thanumalayan Sankaranarayana Pillai and Marie Pellat and Aitor Lewkowycz and Erica Moreira and Rewon Child and Oleksandr Polozov and Katherine Lee and Zongwei Zhou and Xuezhi Wang and Brennan Saeta and Mark Diaz and Orhan Firat and Michele Catasta and Jason Wei and Kathy Meier-Hellstern and Douglas Eck and Jeff Dean and Slav Petrov and Noah Fiedel},
      year={2022},
      eprint={2204.02311},
      archivePrefix={arXiv},
      primaryClass={cs.CL},
      url={https://arxiv.org/abs/2204.02311}, 
}

@inproceedings{lu-etal-2025-paths,
    title = "Paths Not Taken: Understanding and Mending the Multilingual Factual Recall Pipeline",
    author = "Lu, Meng  and
      Zhang, Ruochen  and
      Eickhoff, Carsten  and
      Pavlick, Ellie",
    editor = "Christodoulopoulos, Christos  and
      Chakraborty, Tanmoy  and
      Rose, Carolyn  and
      Peng, Violet",
    booktitle = "Proceedings of the 2025 Conference on Empirical Methods in Natural Language Processing",
    month = nov,
    year = "2025",
    address = "Suzhou, China",
    publisher = "Association for Computational Linguistics",
    url = "https://aclanthology.org/2025.emnlp-main.762/",
    doi = "10.18653/v1/2025.emnlp-main.762",
    pages = "15066--15096",
    ISBN = "979-8-89176-332-6"
}

@inproceedings{chi-etal-2020-finding,
    title = "Finding Universal Grammatical Relations in Multilingual {BERT}",
    author = "Chi, Ethan A.  and
      Hewitt, John  and
      Manning, Christopher D.",
    editor = "Jurafsky, Dan  and
      Chai, Joyce  and
      Schluter, Natalie  and
      Tetreault, Joel",
    booktitle = "Proceedings of the 58th Annual Meeting of the Association for Computational Linguistics",
    month = jul,
    year = "2020",
    address = "Online",
    publisher = "Association for Computational Linguistics",
    url = "https://aclanthology.org/2020.acl-main.493/",
    doi = "10.18653/v1/2020.acl-main.493",
    pages = "5564--5577"
}

@inproceedings{krubinski-pecina-2022-comet,
    title = "From {COMET} to {COMES} {--} Can Summary Evaluation Benefit from Translation Evaluation?",
    author = "Krubi{\'n}ski, Mateusz  and
      Pecina, Pavel",
    editor = "Deutsch, Daniel  and
      Udomcharoenchaikit, Can  and
      Opitz, Juri  and
      Gao, Yang  and
      Fomicheva, Marina  and
      Eger, Steffen",
    booktitle = "Proceedings of the 3rd Workshop on Evaluation and Comparison of NLP Systems",
    month = nov,
    year = "2022",
    address = "Online",
    publisher = "Association for Computational Linguistics",
    url = "https://aclanthology.org/2022.eval4nlp-1.3/",
    doi = "10.18653/v1/2022.eval4nlp-1.3",
    pages = "21--31"
}

@inproceedings{clark-etal-2023-seahorse,
    title = "{SEAHORSE}: A Multilingual, Multifaceted Dataset for Summarization Evaluation",
    author = "Clark, Elizabeth  and
      Rijhwani, Shruti  and
      Gehrmann, Sebastian  and
      Maynez, Joshua  and
      Aharoni, Roee  and
      Nikolaev, Vitaly  and
      Sellam, Thibault  and
      Siddhant, Aditya  and
      Das, Dipanjan  and
      Parikh, Ankur",
    editor = "Bouamor, Houda  and
      Pino, Juan  and
      Bali, Kalika",
    booktitle = "Proceedings of the 2023 Conference on Empirical Methods in Natural Language Processing",
    month = dec,
    year = "2023",
    address = "Singapore",
    publisher = "Association for Computational Linguistics",
    url = "https://aclanthology.org/2023.emnlp-main.584/",
    doi = "10.18653/v1/2023.emnlp-main.584",
    pages = "9397--9413"
}

@inproceedings{conneau-etal-2020-unsupervised,
    title = "Unsupervised Cross-lingual Representation Learning at Scale",
    author = "Conneau, Alexis  and
      Khandelwal, Kartikay  and
      Goyal, Naman  and
      Chaudhary, Vishrav  and
      Wenzek, Guillaume  and
      Guzm{\'a}n, Francisco  and
      Grave, Edouard  and
      Ott, Myle  and
      Zettlemoyer, Luke  and
      Stoyanov, Veselin",
    editor = "Jurafsky, Dan  and
      Chai, Joyce  and
      Schluter, Natalie  and
      Tetreault, Joel",
    booktitle = "Proceedings of the 58th Annual Meeting of the Association for Computational Linguistics",
    month = jul,
    year = "2020",
    address = "Online",
    publisher = "Association for Computational Linguistics",
    url = "https://aclanthology.org/2020.acl-main.747/",
    doi = "10.18653/v1/2020.acl-main.747",
    pages = "8440--8451"
}

@inproceedings{hasan-etal-2021-xl,
    title = "{XL}-Sum: Large-Scale Multilingual Abstractive Summarization for 44 Languages",
    author = "Hasan, Tahmid  and
      Bhattacharjee, Abhik  and
      Islam, Md. Saiful  and
      Mubasshir, Kazi  and
      Li, Yuan-Fang  and
      Kang, Yong-Bin  and
      Rahman, M. Sohel  and
      Shahriyar, Rifat",
    editor = "Zong, Chengqing  and
      Xia, Fei  and
      Li, Wenjie  and
      Navigli, Roberto",
    booktitle = "Findings of the Association for Computational Linguistics: ACL-IJCNLP 2021",
    month = aug,
    year = "2021",
    address = "Online",
    publisher = "Association for Computational Linguistics",
    url = "https://aclanthology.org/2021.findings-acl.413/",
    doi = "10.18653/v1/2021.findings-acl.413",
    pages = "4693--4703"
}

@article{flores,
    title = "The {F}lores-101 Evaluation Benchmark for Low-Resource and Multilingual Machine Translation",
    author = "Goyal, Naman  and
      Gao, Cynthia  and
      Chaudhary, Vishrav  and
      Chen, Peng-Jen  and
      Wenzek, Guillaume  and
      Ju, Da  and
      Krishnan, Sanjana  and
      Ranzato, Marc{'}Aurelio  and
      Guzm{\'a}n, Francisco  and
      Fan, Angela",
    editor = "Roark, Brian  and
      Nenkova, Ani",
    journal = "Transactions of the Association for Computational Linguistics",
    volume = "10",
    year = "2022",
    address = "Cambridge, MA",
    publisher = "MIT Press",
    url = "https://aclanthology.org/2022.tacl-1.30/",
    doi = "10.1162/tacl_a_00474",
    pages = "522--538"
}

@article{fabbri-etal-2021-summeval,
    title = "{S}umm{E}val: Re-evaluating Summarization Evaluation",
    author = "Fabbri, Alexander R.  and
      Kry{\'s}ci{\'n}ski, Wojciech  and
      McCann, Bryan  and
      Xiong, Caiming  and
      Socher, Richard  and
      Radev, Dragomir",
    editor = "Roark, Brian  and
      Nenkova, Ani",
    journal = "Transactions of the Association for Computational Linguistics",
    volume = "9",
    year = "2021",
    address = "Cambridge, MA",
    publisher = "MIT Press",
    url = "https://aclanthology.org/2021.tacl-1.24/",
    doi = "10.1162/tacl_a_00373",
    pages = "391--409"
}

@inproceedings{wendler-etal-2024-llamas,
    title = "Do Llamas Work in {E}nglish? On the Latent Language of Multilingual Transformers",
    author = "Wendler, Chris  and
      Veselovsky, Veniamin  and
      Monea, Giovanni  and
      West, Robert",
    editor = "Ku, Lun-Wei  and
      Martins, Andre  and
      Srikumar, Vivek",
    booktitle = "Proceedings of the 62nd Annual Meeting of the Association for Computational Linguistics (Volume 1: Long Papers)",
    month = aug,
    year = "2024",
    address = "Bangkok, Thailand",
    publisher = "Association for Computational Linguistics",
    url = "https://aclanthology.org/2024.acl-long.820/",
    doi = "10.18653/v1/2024.acl-long.820",
    pages = "15366--15394"
}

@inproceedings{fu-etal-2024-gptscore,
    title = "{GPTS}core: Evaluate as You Desire",
    author = "Fu, Jinlan  and
      Ng, See-Kiong  and
      Jiang, Zhengbao  and
      Liu, Pengfei",
    editor = "Duh, Kevin  and
      Gomez, Helena  and
      Bethard, Steven",
    booktitle = "Proceedings of the 2024 Conference of the North American Chapter of the Association for Computational Linguistics: Human Language Technologies (Volume 1: Long Papers)",
    month = jun,
    year = "2024",
    address = "Mexico City, Mexico",
    publisher = "Association for Computational Linguistics",
    url = "https://aclanthology.org/2024.naacl-long.365/",
    doi = "10.18653/v1/2024.naacl-long.365",
    pages = "6556--6576"
}

@inproceedings{liu-etal-2023-g,
    title = "{G}-Eval: {NLG} Evaluation using Gpt-4 with Better Human Alignment",
    author = "Liu, Yang  and
      Iter, Dan  and
      Xu, Yichong  and
      Wang, Shuohang  and
      Xu, Ruochen  and
      Zhu, Chenguang",
    editor = "Bouamor, Houda  and
      Pino, Juan  and
      Bali, Kalika",
    booktitle = "Proceedings of the 2023 Conference on Empirical Methods in Natural Language Processing",
    month = dec,
    year = "2023",
    address = "Singapore",
    publisher = "Association for Computational Linguistics",
    url = "https://aclanthology.org/2023.emnlp-main.153/",
    doi = "10.18653/v1/2023.emnlp-main.153",
    pages = "2511--2522"
}

@inproceedings{sellam-etal-2020-bleurt,
    title = "{BLEURT}: Learning Robust Metrics for Text Generation",
    author = "Sellam, Thibault  and
      Das, Dipanjan  and
      Parikh, Ankur",
    editor = "Jurafsky, Dan  and
      Chai, Joyce  and
      Schluter, Natalie  and
      Tetreault, Joel",
    booktitle = "Proceedings of the 58th Annual Meeting of the Association for Computational Linguistics",
    month = jul,
    year = "2020",
    address = "Online",
    publisher = "Association for Computational Linguistics",
    url = "https://aclanthology.org/2020.acl-main.704/",
    doi = "10.18653/v1/2020.acl-main.704",
    pages = "7881--7892"
}

@inproceedings{rei-etal-2020-comet,
    title = "{COMET}: A Neural Framework for {MT} Evaluation",
    author = "Rei, Ricardo  and
      Stewart, Craig  and
      Farinha, Ana C  and
      Lavie, Alon",
    editor = "Webber, Bonnie  and
      Cohn, Trevor  and
      He, Yulan  and
      Liu, Yang",
    booktitle = "Proceedings of the 2020 Conference on Empirical Methods in Natural Language Processing (EMNLP)",
    month = nov,
    year = "2020",
    address = "Online",
    publisher = "Association for Computational Linguistics",
    url = "https://aclanthology.org/2020.emnlp-main.213/",
    doi = "10.18653/v1/2020.emnlp-main.213",
    pages = "2685--2702"
}

@inproceedings{sheng-etal-2024-repeval,
    title = "{R}ep{E}val: Effective Text Evaluation with {LLM} Representation",
    author = "Sheng, Shuqian  and
      Xu, Yi  and
      Zhang, Tianhang  and
      Shen, Zanwei  and
      Fu, Luoyi  and
      Ding, Jiaxin  and
      Zhou, Lei  and
      Gan, Xiaoying  and
      Wang, Xinbing  and
      Zhou, Chenghu",
    editor = "Al-Onaizan, Yaser  and
      Bansal, Mohit  and
      Chen, Yun-Nung",
    booktitle = "Proceedings of the 2024 Conference on Empirical Methods in Natural Language Processing",
    month = nov,
    year = "2024",
    address = "Miami, Florida, USA",
    publisher = "Association for Computational Linguistics",
    url = "https://aclanthology.org/2024.emnlp-main.398/",
    doi = "10.18653/v1/2024.emnlp-main.398",
    pages = "7019--7033"
}

@inproceedings{hada-etal-2024-metal,
    title = "{METAL}: Towards Multilingual Meta-Evaluation",
    author = "Hada, Rishav  and
      Gumma, Varun  and
      Ahmed, Mohamed  and
      Bali, Kalika  and
      Sitaram, Sunayana",
    editor = "Duh, Kevin  and
      Gomez, Helena  and
      Bethard, Steven",
    booktitle = "Findings of the Association for Computational Linguistics: NAACL 2024",
    month = jun,
    year = "2024",
    address = "Mexico City, Mexico",
    publisher = "Association for Computational Linguistics",
    url = "https://aclanthology.org/2024.findings-naacl.148/",
    doi = "10.18653/v1/2024.findings-naacl.148",
    pages = "2280--2298"
}

@inproceedings{dumas-etal-2025-separating,
    title = "Separating Tongue from Thought: Activation Patching Reveals Language-Agnostic Concept Representations in Transformers",
    author = "Dumas, Cl{\'e}ment  and
      Wendler, Chris  and
      Veselovsky, Veniamin  and
      Monea, Giovanni  and
      West, Robert",
    editor = "Che, Wanxiang  and
      Nabende, Joyce  and
      Shutova, Ekaterina  and
      Pilehvar, Mohammad Taher",
    booktitle = "Proceedings of the 63rd Annual Meeting of the Association for Computational Linguistics (Volume 1: Long Papers)",
    month = jul,
    year = "2025",
    address = "Vienna, Austria",
    publisher = "Association for Computational Linguistics",
    url = "https://aclanthology.org/2025.acl-long.1536/",
    doi = "10.18653/v1/2025.acl-long.1536",
    pages = "31822--31841",
    ISBN = "979-8-89176-251-0"
}

@misc{wang2025refusaldirectionuniversalsafetyaligned,
      title={Refusal Direction is Universal Across Safety-Aligned Languages}, 
      author={Xinpeng Wang and Mingyang Wang and Yihong Liu and Hinrich Schütze and Barbara Plank},
      year={2025},
      eprint={2505.17306},
      archivePrefix={arXiv},
      primaryClass={cs.CL},
      url={https://arxiv.org/abs/2505.17306}, 
}

@inproceedings{sterz-etal-2025-recover,
    title = "{R}e{C}o{V}e{R} the Target Language: Language Steering without Sacrificing Task Performance",
    author = "Sterz, Hannah  and
      Schmidt, Fabian David  and
      Glava{\v{s}}, Goran  and
      Vuli{\'c}, Ivan",
    editor = "Christodoulopoulos, Christos  and
      Chakraborty, Tanmoy  and
      Rose, Carolyn  and
      Peng, Violet",
    booktitle = "Findings of the Association for Computational Linguistics: EMNLP 2025",
    month = nov,
    year = "2025",
    address = "Suzhou, China",
    publisher = "Association for Computational Linguistics",
    url = "https://aclanthology.org/2025.findings-emnlp.1056/",
    doi = "10.18653/v1/2025.findings-emnlp.1056",
    pages = "19390--19405",
    ISBN = "979-8-89176-335-7"
}

@misc{arditi2024refusallanguagemodelsmediated,
      title={Refusal in Language Models Is Mediated by a Single Direction}, 
      author={Andy Arditi and Oscar Obeso and Aaquib Syed and Daniel Paleka and Nina Panickssery and Wes Gurnee and Neel Nanda},
      year={2024},
      eprint={2406.11717},
      archivePrefix={arXiv},
      primaryClass={cs.LG},
      url={https://arxiv.org/abs/2406.11717}, 
}

@misc{marks2024geometrytruthemergentlinear,
      title={The Geometry of Truth: Emergent Linear Structure in Large Language Model Representations of True/False Datasets}, 
      author={Samuel Marks and Max Tegmark},
      year={2024},
      eprint={2310.06824},
      archivePrefix={arXiv},
      primaryClass={cs.AI},
      url={https://arxiv.org/abs/2310.06824}, 
}

@misc{turner2024steeringlanguagemodelsactivation,
      title={Steering Language Models With Activation Engineering}, 
      author={Alexander Matt Turner and Lisa Thiergart and Gavin Leech and David Udell and Juan J. Vazquez and Ulisse Mini and Monte MacDiarmid},
      year={2024},
      eprint={2308.10248},
      archivePrefix={arXiv},
      primaryClass={cs.CL},
      url={https://arxiv.org/abs/2308.10248}, 
}

@inproceedings{10.5555/3692070.3693675,
author = {Park, Kiho and Choe, Yo Joong and Veitch, Victor},
title = {The linear representation hypothesis and the geometry of large language models},
year = {2024},
publisher = {JMLR.org},
booktitle = {Proceedings of the 41st International Conference on Machine Learning},
articleno = {1605},
numpages = {24},
location = {Vienna, Austria},
series = {ICML'24}
}

@misc{bafna2025translationbarrierhypothesismultilingual,
      title={The Translation Barrier Hypothesis: Multilingual Generation with Large Language Models Suffers from Implicit Translation Failure}, 
      author={Niyati Bafna and Tianjian Li and Kenton Murray and David R. Mortensen and David Yarowsky and Hale Sirin and Daniel Khashabi},
      year={2025},
      eprint={2506.22724},
      archivePrefix={arXiv},
      primaryClass={cs.CL},
      url={https://arxiv.org/abs/2506.22724}, 
}

@inproceedings{bavaresco-etal-2025-llms,
    title = "{LLM}s instead of Human Judges? A Large Scale Empirical Study across 20 {NLP} Evaluation Tasks",
    author = "Bavaresco, Anna  and
      Bernardi, Raffaella  and
      Bertolazzi, Leonardo  and
      Elliott, Desmond  and
      Fern{\'a}ndez, Raquel  and
      Gatt, Albert  and
      Ghaleb, Esam  and
      Giulianelli, Mario  and
      Hanna, Michael  and
      Koller, Alexander  and
      Martins, Andre  and
      Mondorf, Philipp  and
      Neplenbroek, Vera  and
      Pezzelle, Sandro  and
      Plank, Barbara  and
      Schlangen, David  and
      Suglia, Alessandro  and
      Surikuchi, Aditya K  and
      Takmaz, Ece  and
      Testoni, Alberto",
    editor = "Che, Wanxiang  and
      Nabende, Joyce  and
      Shutova, Ekaterina  and
      Pilehvar, Mohammad Taher",
    booktitle = "Proceedings of the 63rd Annual Meeting of the Association for Computational Linguistics (Volume 2: Short Papers)",
    month = jul,
    year = "2025",
    address = "Vienna, Austria",
    publisher = "Association for Computational Linguistics",
    url = "https://aclanthology.org/2025.acl-short.20/",
    doi = "10.18653/v1/2025.acl-short.20",
    pages = "238--255",
    ISBN = "979-8-89176-252-7"
}

@inproceedings{casola-etal-2025-references,
    title = "References Matter: Investigating the Impact of Reference Set Variation on Summarization Evaluation",
    author = "Casola, Silvia  and
      Liu, Yang Janet  and
      Peng, Siyao  and
      Kraus, Oliver  and
      Gatt, Albert  and
      Plank, Barbara",
    editor = "Flek, Lucie  and
      Narayan, Shashi  and
      Phương, L{\^e} Hồng  and
      Pei, Jiahuan",
    booktitle = "Proceedings of the 18th International Natural Language Generation Conference",
    month = oct,
    year = "2025",
    address = "Hanoi, Vietnam",
    publisher = "Association for Computational Linguistics",
    url = "https://aclanthology.org/2025.inlg-main.18/",
    pages = "274--291"
}

@inproceedings{
Zhang*2020BERTScore,
title={BERTScore: Evaluating Text Generation with BERT},
author={Tianyi Zhang and Varsha Kishore and Felix Wu and Kilian Q. Weinberger and Yoav Artzi},
booktitle={International Conference on Learning Representations},
year={2020},
url={https://openreview.net/forum?id=SkeHuCVFDr}
}

@misc{celikyilmaz2021evaluationtextgenerationsurvey,
      title={Evaluation of Text Generation: A Survey}, 
      author={Asli Celikyilmaz and Elizabeth Clark and Jianfeng Gao},
      year={2021},
      eprint={2006.14799},
      archivePrefix={arXiv},
      primaryClass={cs.CL},
      url={https://arxiv.org/abs/2006.14799}, 
}

@article{10.1145/3485766,
author = {Sai, Ananya B. and Mohankumar, Akash Kumar and Khapra, Mitesh M.},
title = {A Survey of Evaluation Metrics Used for NLG Systems},
year = {2022},
issue_date = {February 2023},
publisher = {Association for Computing Machinery},
address = {New York, NY, USA},
volume = {55},
number = {2},
issn = {0360-0300},
url = {https://doi.org/10.1145/3485766},
doi = {10.1145/3485766},
journal = {ACM Comput. Surv.},
month = jan,
articleno = {26},
numpages = {39}
}

@inproceedings{wang-etal-2025-bridging,
    title = "Bridging the Language Gaps in Large Language Models with Inference-Time Cross-Lingual Intervention",
    author = "Wang, Weixuan  and
      Wu, Minghao  and
      Haddow, Barry  and
      Birch, Alexandra",
    editor = "Che, Wanxiang  and
      Nabende, Joyce  and
      Shutova, Ekaterina  and
      Pilehvar, Mohammad Taher",
    booktitle = "Proceedings of the 63rd Annual Meeting of the Association for Computational Linguistics (Volume 1: Long Papers)",
    month = jul,
    year = "2025",
    address = "Vienna, Austria",
    publisher = "Association for Computational Linguistics",
    url = "https://aclanthology.org/2025.acl-long.270/",
    doi = "10.18653/v1/2025.acl-long.270",
    pages = "5418--5433",
    ISBN = "979-8-89176-251-0"
}

@misc{dang2024ayaexpansecombiningresearch,
      title={Aya Expanse: Combining Research Breakthroughs for a New Multilingual Frontier}, 
      author={John Dang and Shivalika Singh and Daniel D'souza and Arash Ahmadian and Alejandro Salamanca and Madeline Smith and Aidan Peppin and Sungjin Hong and Manoj Govindassamy and Terrence Zhao and Sandra Kublik and Meor Amer and Viraat Aryabumi and Jon Ander Campos and Yi-Chern Tan and Tom Kocmi and Florian Strub and Nathan Grinsztajn and Yannis Flet-Berliac and Acyr Locatelli and Hangyu Lin and Dwarak Talupuru and Bharat Venkitesh and David Cairuz and Bowen Yang and Tim Chung and Wei-Yin Ko and Sylvie Shang Shi and Amir Shukayev and Sammie Bae and Aleksandra Piktus and Roman Castagné and Felipe Cruz-Salinas and Eddie Kim and Lucas Crawhall-Stein and Adrien Morisot and Sudip Roy and Phil Blunsom and Ivan Zhang and Aidan Gomez and Nick Frosst and Marzieh Fadaee and Beyza Ermis and Ahmet Üstün and Sara Hooker},
      year={2024},
      eprint={2412.04261},
      archivePrefix={arXiv},
      primaryClass={cs.CL},
      url={https://arxiv.org/abs/2412.04261}, 
}

@misc{grattafiori2024llama3herdmodels,
      title={The Llama 3 Herd of Models}, 
      author={Aaron Grattafiori and Abhimanyu Dubey and Abhinav Jauhri and Abhinav Pandey and Abhishek Kadian and Ahmad Al-Dahle and Aiesha Letman and Akhil Mathur and Alan Schelten and Alex Vaughan and Amy Yang and Angela Fan and Anirudh Goyal and Anthony Hartshorn and Aobo Yang and Archi Mitra and Archie Sravankumar and Artem Korenev and Arthur Hinsvark and Arun Rao and Aston Zhang and Aurelien Rodriguez and Austen Gregerson and Ava Spataru and Baptiste Roziere and Bethany Biron and Binh Tang and Bobbie Chern and Charlotte Caucheteux and Chaya Nayak and Chloe Bi and Chris Marra and Chris McConnell and Christian Keller and Christophe Touret and Chunyang Wu and Corinne Wong and Cristian Canton Ferrer and Cyrus Nikolaidis and Damien Allonsius and Daniel Song and Danielle Pintz and Danny Livshits and Danny Wyatt and David Esiobu and Dhruv Choudhary and Dhruv Mahajan and Diego Garcia-Olano and Diego Perino and Dieuwke Hupkes and Egor Lakomkin and Ehab AlBadawy and Elina Lobanova and Emily Dinan and Eric Michael Smith and Filip Radenovic and Francisco Guzmán and Frank Zhang and Gabriel Synnaeve and Gabrielle Lee and Georgia Lewis Anderson and Govind Thattai and Graeme Nail and Gregoire Mialon and Guan Pang and Guillem Cucurell and Hailey Nguyen and Hannah Korevaar and Hu Xu and Hugo Touvron and Iliyan Zarov and Imanol Arrieta Ibarra and Isabel Kloumann and Ishan Misra and Ivan Evtimov and Jack Zhang and Jade Copet and Jaewon Lee and Jan Geffert and Jana Vranes and Jason Park and Jay Mahadeokar and Jeet Shah and Jelmer van der Linde and Jennifer Billock and Jenny Hong and Jenya Lee and Jeremy Fu and Jianfeng Chi and Jianyu Huang and Jiawen Liu and Jie Wang and Jiecao Yu and Joanna Bitton and Joe Spisak and Jongsoo Park and Joseph Rocca and Joshua Johnstun and Joshua Saxe and Junteng Jia and Kalyan Vasuden Alwala and Karthik Prasad and Kartikeya Upasani and Kate Plawiak and Ke Li and Kenneth Heafield and Kevin Stone and Khalid El-Arini and Krithika Iyer and Kshitiz Malik and Kuenley Chiu and Kunal Bhalla and Kushal Lakhotia and Lauren Rantala-Yeary and Laurens van der Maaten and Lawrence Chen and Liang Tan and Liz Jenkins and Louis Martin and Lovish Madaan and Lubo Malo and Lukas Blecher and Lukas Landzaat and Luke de Oliveira and Madeline Muzzi and Mahesh Pasupuleti and Mannat Singh and Manohar Paluri and Marcin Kardas and Maria Tsimpoukelli and Mathew Oldham and Mathieu Rita and Maya Pavlova and Melanie Kambadur and Mike Lewis and Min Si and Mitesh Kumar Singh and Mona Hassan and Naman Goyal and Narjes Torabi and Nikolay Bashlykov and Nikolay Bogoychev and Niladri Chatterji and Ning Zhang and Olivier Duchenne and Onur Çelebi and Patrick Alrassy and Pengchuan Zhang and Pengwei Li and Petar Vasic and Peter Weng and Prajjwal Bhargava and Pratik Dubal and Praveen Krishnan and Punit Singh Koura and Puxin Xu and Qing He and Qingxiao Dong and Ragavan Srinivasan and Raj Ganapathy and Ramon Calderer and Ricardo Silveira Cabral and Robert Stojnic and Roberta Raileanu and Rohan Maheswari and Rohit Girdhar and Rohit Patel and Romain Sauvestre and Ronnie Polidoro and Roshan Sumbaly and Ross Taylor and Ruan Silva and Rui Hou and Rui Wang and Saghar Hosseini and Sahana Chennabasappa and Sanjay Singh and Sean Bell and Seohyun Sonia Kim and Sergey Edunov and Shaoliang Nie and Sharan Narang and Sharath Raparthy and Sheng Shen and Shengye Wan and Shruti Bhosale and Shun Zhang and Simon Vandenhende and Soumya Batra and Spencer Whitman and Sten Sootla and Stephane Collot and Suchin Gururangan and Sydney Borodinsky and Tamar Herman and Tara Fowler and Tarek Sheasha and Thomas Georgiou and Thomas Scialom and Tobias Speckbacher and Todor Mihaylov and Tong Xiao and Ujjwal Karn and Vedanuj Goswami and Vibhor Gupta and Vignesh Ramanathan and Viktor Kerkez and Vincent Gonguet and Virginie Do and Vish Vogeti and Vítor Albiero and Vladan Petrovic and Weiwei Chu and Wenhan Xiong and Wenyin Fu and Whitney Meers and Xavier Martinet and Xiaodong Wang and Xiaofang Wang and Xiaoqing Ellen Tan and Xide Xia and Xinfeng Xie and Xuchao Jia and Xuewei Wang and Yaelle Goldschlag and Yashesh Gaur and Yasmine Babaei and Yi Wen and Yiwen Song and Yuchen Zhang and Yue Li and Yuning Mao and Zacharie Delpierre Coudert and Zheng Yan and Zhengxing Chen and Zoe Papakipos and Aaditya Singh and Aayushi Srivastava and Abha Jain and Adam Kelsey and Adam Shajnfeld and Adithya Gangidi and Adolfo Victoria and Ahuva Goldstand and Ajay Menon and Ajay Sharma and Alex Boesenberg and Alexei Baevski and Allie Feinstein and Amanda Kallet and Amit Sangani and Amos Teo and Anam Yunus and Andrei Lupu and Andres Alvarado and Andrew Caples and Andrew Gu and Andrew Ho and Andrew Poulton and Andrew Ryan and Ankit Ramchandani and Annie Dong and Annie Franco and Anuj Goyal and Aparajita Saraf and Arkabandhu Chowdhury and Ashley Gabriel and Ashwin Bharambe and Assaf Eisenman and Azadeh Yazdan and Beau James and Ben Maurer and Benjamin Leonhardi and Bernie Huang and Beth Loyd and Beto De Paola and Bhargavi Paranjape and Bing Liu and Bo Wu and Boyu Ni and Braden Hancock and Bram Wasti and Brandon Spence and Brani Stojkovic and Brian Gamido and Britt Montalvo and Carl Parker and Carly Burton and Catalina Mejia and Ce Liu and Changhan Wang and Changkyu Kim and Chao Zhou and Chester Hu and Ching-Hsiang Chu and Chris Cai and Chris Tindal and Christoph Feichtenhofer and Cynthia Gao and Damon Civin and Dana Beaty and Daniel Kreymer and Daniel Li and David Adkins and David Xu and Davide Testuggine and Delia David and Devi Parikh and Diana Liskovich and Didem Foss and Dingkang Wang and Duc Le and Dustin Holland and Edward Dowling and Eissa Jamil and Elaine Montgomery and Eleonora Presani and Emily Hahn and Emily Wood and Eric-Tuan Le and Erik Brinkman and Esteban Arcaute and Evan Dunbar and Evan Smothers and Fei Sun and Felix Kreuk and Feng Tian and Filippos Kokkinos and Firat Ozgenel and Francesco Caggioni and Frank Kanayet and Frank Seide and Gabriela Medina Florez and Gabriella Schwarz and Gada Badeer and Georgia Swee and Gil Halpern and Grant Herman and Grigory Sizov and Guangyi and Zhang and Guna Lakshminarayanan and Hakan Inan and Hamid Shojanazeri and Han Zou and Hannah Wang and Hanwen Zha and Haroun Habeeb and Harrison Rudolph and Helen Suk and Henry Aspegren and Hunter Goldman and Hongyuan Zhan and Ibrahim Damlaj and Igor Molybog and Igor Tufanov and Ilias Leontiadis and Irina-Elena Veliche and Itai Gat and Jake Weissman and James Geboski and James Kohli and Janice Lam and Japhet Asher and Jean-Baptiste Gaya and Jeff Marcus and Jeff Tang and Jennifer Chan and Jenny Zhen and Jeremy Reizenstein and Jeremy Teboul and Jessica Zhong and Jian Jin and Jingyi Yang and Joe Cummings and Jon Carvill and Jon Shepard and Jonathan McPhie and Jonathan Torres and Josh Ginsburg and Junjie Wang and Kai Wu and Kam Hou U and Karan Saxena and Kartikay Khandelwal and Katayoun Zand and Kathy Matosich and Kaushik Veeraraghavan and Kelly Michelena and Keqian Li and Kiran Jagadeesh and Kun Huang and Kunal Chawla and Kyle Huang and Lailin Chen and Lakshya Garg and Lavender A and Leandro Silva and Lee Bell and Lei Zhang and Liangpeng Guo and Licheng Yu and Liron Moshkovich and Luca Wehrstedt and Madian Khabsa and Manav Avalani and Manish Bhatt and Martynas Mankus and Matan Hasson and Matthew Lennie and Matthias Reso and Maxim Groshev and Maxim Naumov and Maya Lathi and Meghan Keneally and Miao Liu and Michael L. Seltzer and Michal Valko and Michelle Restrepo and Mihir Patel and Mik Vyatskov and Mikayel Samvelyan and Mike Clark and Mike Macey and Mike Wang and Miquel Jubert Hermoso and Mo Metanat and Mohammad Rastegari and Munish Bansal and Nandhini Santhanam and Natascha Parks and Natasha White and Navyata Bawa and Nayan Singhal and Nick Egebo and Nicolas Usunier and Nikhil Mehta and Nikolay Pavlovich Laptev and Ning Dong and Norman Cheng and Oleg Chernoguz and Olivia Hart and Omkar Salpekar and Ozlem Kalinli and Parkin Kent and Parth Parekh and Paul Saab and Pavan Balaji and Pedro Rittner and Philip Bontrager and Pierre Roux and Piotr Dollar and Polina Zvyagina and Prashant Ratanchandani and Pritish Yuvraj and Qian Liang and Rachad Alao and Rachel Rodriguez and Rafi Ayub and Raghotham Murthy and Raghu Nayani and Rahul Mitra and Rangaprabhu Parthasarathy and Raymond Li and Rebekkah Hogan and Robin Battey and Rocky Wang and Russ Howes and Ruty Rinott and Sachin Mehta and Sachin Siby and Sai Jayesh Bondu and Samyak Datta and Sara Chugh and Sara Hunt and Sargun Dhillon and Sasha Sidorov and Satadru Pan and Saurabh Mahajan and Saurabh Verma and Seiji Yamamoto and Sharadh Ramaswamy and Shaun Lindsay and Shaun Lindsay and Sheng Feng and Shenghao Lin and Shengxin Cindy Zha and Shishir Patil and Shiva Shankar and Shuqiang Zhang and Shuqiang Zhang and Sinong Wang and Sneha Agarwal and Soji Sajuyigbe and Soumith Chintala and Stephanie Max and Stephen Chen and Steve Kehoe and Steve Satterfield and Sudarshan Govindaprasad and Sumit Gupta and Summer Deng and Sungmin Cho and Sunny Virk and Suraj Subramanian and Sy Choudhury and Sydney Goldman and Tal Remez and Tamar Glaser and Tamara Best and Thilo Koehler and Thomas Robinson and Tianhe Li and Tianjun Zhang and Tim Matthews and Timothy Chou and Tzook Shaked and Varun Vontimitta and Victoria Ajayi and Victoria Montanez and Vijai Mohan and Vinay Satish Kumar and Vishal Mangla and Vlad Ionescu and Vlad Poenaru and Vlad Tiberiu Mihailescu and Vladimir Ivanov and Wei Li and Wenchen Wang and Wenwen Jiang and Wes Bouaziz and Will Constable and Xiaocheng Tang and Xiaojian Wu and Xiaolan Wang and Xilun Wu and Xinbo Gao and Yaniv Kleinman and Yanjun Chen and Ye Hu and Ye Jia and Ye Qi and Yenda Li and Yilin Zhang and Ying Zhang and Yossi Adi and Youngjin Nam and Yu and Wang and Yu Zhao and Yuchen Hao and Yundi Qian and Yunlu Li and Yuzi He and Zach Rait and Zachary DeVito and Zef Rosnbrick and Zhaoduo Wen and Zhenyu Yang and Zhiwei Zhao and Zhiyu Ma},
      year={2024},
      eprint={2407.21783},
      archivePrefix={arXiv},
      primaryClass={cs.AI},
      url={https://arxiv.org/abs/2407.21783}, 
}

@inproceedings{mondshine-etal-2025-beyond-n,
    title = "Beyond N-Grams: Rethinking Evaluation Metrics and Strategies for Multilingual Abstractive Summarization",
    author = "Mondshine, Itai  and
      Paz-Argaman, Tzuf  and
      Tsarfaty, Reut",
    editor = "Che, Wanxiang  and
      Nabende, Joyce  and
      Shutova, Ekaterina  and
      Pilehvar, Mohammad Taher",
    booktitle = "Proceedings of the 63rd Annual Meeting of the Association for Computational Linguistics (Volume 1: Long Papers)",
    month = jul,
    year = "2025",
    address = "Vienna, Austria",
    publisher = "Association for Computational Linguistics",
    url = "https://aclanthology.org/2025.acl-long.932/",
    doi = "10.18653/v1/2025.acl-long.932",
    pages = "19019--19035",
    ISBN = "979-8-89176-251-0"
}

@misc{barnes2025summarizationmetricsspanishbasque,
      title={Summarization Metrics for Spanish and Basque: Do Automatic Scores and LLM-Judges Correlate with Humans?}, 
      author={Jeremy Barnes and Naiara Perez and Alba Bonet-Jover and Begoña Altuna},
      year={2025},
      eprint={2503.17039},
      archivePrefix={arXiv},
      primaryClass={cs.CL},
      url={https://arxiv.org/abs/2503.17039}, 
}

\clearpage

\appendix

\section{Dimension-specific Results}
\label{app:dimensions}
Tables \ref{tab:seahorse_perdim_question1_en}, \ref{tab:seahorse_perdim_question2_en}, \ref{tab:seahorse_perdim_question3_en}, 
\ref{tab:seahorse_perdim_question4_en}, \ref{tab:seahorse_perdim_question5_en}, and \ref{tab:seahorse_perdim_question6_en} present the 
results for the six evaluation dimensions in Seahorse when using an English prompt. 

\begin{table*}[]
\centering
\resizebox{\textwidth}{!}{%
\begin{tabular}{ll|c|c|c|c|c}
\textbf{} & \textbf{} & \textbf{de} & \textbf{es} & \textbf{ru} & \textbf{tr} & \textbf{vi} \\
\toprule
\multicolumn{7}{l}{\textbf{GPTScore}} \\
\midrule
\multirow{3}{*}{\texttt{Llama-3.2-3B}} & Baseline (EN-prompt) & $.003$ & $.319$ & $.313$ & $.099$ & $.248$ \\
 & + vector steer (EN-prompt) & \textcolor{blue!70!black}{$\underline{.025}$$\uparrow$} & \textcolor{blue!70!black}{$.320$$\uparrow$} & 
\textcolor{blue!70!black}{$.315$$\uparrow$} & \textcolor{blue!70!black}{$\underline{.158}$$\uparrow$} & 
\textcolor{red!70!black}{$.244$$\downarrow$} \\
 & + avgvec zero-shot (EN-prompt) & \textcolor{blue!70!black}{$.023$$\uparrow$} & \textcolor{blue!70!black}{$\underline{.333}$$\uparrow$} 
& \textcolor{blue!70!black}{$\underline{.331}$$\uparrow$} & \textcolor{blue!70!black}{$.140$$\uparrow$} & 
\textcolor{blue!70!black}{$\underline{.258}$$\uparrow$} \\
\midrule
\multirow{3}{*}{\texttt{aya-expanse-8b}} & Baseline (EN-prompt) & $\underline{.029}$ & $.249$ & $.353$ & $.114$ & $.305$ \\
 & + vector steer (EN-prompt) & \textcolor{red!70!black}{$.027$$\downarrow$} & \textcolor{blue!70!black}{$.255$$\uparrow$} & 
\textcolor{blue!70!black}{$.361$$\uparrow$} & \textcolor{blue!70!black}{$\underline{.118}$$\uparrow$} & 
\textcolor{blue!70!black}{$.308$$\uparrow$} \\
 & + avgvec zero-shot (EN-prompt) & \textcolor{red!70!black}{$.029$$\downarrow$} & 
\textcolor{blue!70!black}{$\underline{.256}$$\uparrow$} & \textcolor{blue!70!black}{$\underline{.362}$$\uparrow$} & 
\textcolor{blue!70!black}{$\underline{.118}$$\uparrow$} & \textcolor{blue!70!black}{$\underline{.311}$$\uparrow$} \\
\midrule
\multicolumn{7}{l}{\textbf{Direct Prompting}} \\
\midrule
\multirow{3}{*}{\texttt{Llama-3.2-3B}} & Baseline (EN-prompt) & $\underline{.095}$ & $\underline{-.001}$ & $.055$ & $.056$ & $.044$ \\
 & + vector steer (EN-prompt) & \textcolor{blue!70!black}{$\underline{.095}$$\uparrow$} & \textcolor{red!70!black}{$-.009$$\downarrow$} & 
\textcolor{blue!70!black}{$\underline{.058}$$\uparrow$} & \textcolor{blue!70!black}{$\underline{.067}$$\uparrow$} & 
\textcolor{blue!70!black}{$\underline{.049}$$\uparrow$} \\
 & + avgvec zero-shot (EN-prompt) & --- & --- & --- & --- & --- \\
\midrule
\multirow{3}{*}{\texttt{aya-expanse-8b}} & Baseline (EN-prompt) & $\underline{.207}$ & $.335$ & $\underline{.386}$ & $.477$ & 
$\underline{.338}$ \\
 & + vector steer (EN-prompt) & \textcolor{red!70!black}{$\underline{.206}$$\downarrow$} & 
\textcolor{blue!70!black}{$\underline{.335}$$\uparrow$} & \textcolor{red!70!black}{$.385$$\downarrow$} & 
\textcolor{blue!70!black}{$\underline{.481}$$\uparrow$} & \textcolor{red!70!black}{$.337$$\downarrow$} \\
 & + avgvec zero-shot (EN-prompt) & --- & --- & --- & --- & --- \\
\midrule
\multicolumn{7}{l}{\textbf{COMET}} \\
\midrule
\multirow{3}{*}{\texttt{wmt22-comet-da}} & Baseline & $.297$ & $.421$ & $.578$ & $.651$ & $.425$ \\
 & + vector steer & \textcolor{blue!70!black}{$\mathbf{\underline{.361}}$$\uparrow$} & 
\textcolor{blue!70!black}{$\mathbf{\underline{.555}}$$\uparrow$} & \textcolor{blue!70!black}{$.622$$\uparrow$} & 
\textcolor{red!70!black}{$.626$$\downarrow$} & \textcolor{blue!70!black}{$\mathbf{\underline{.454}}$$\uparrow$} \\
 & + avgvec zero-shot & \textcolor{blue!70!black}{$.346$$\uparrow$} & \textcolor{blue!70!black}{$.489$$\uparrow$} & 
\textcolor{blue!70!black}{$\mathbf{\underline{.627}}$$\uparrow$} & \textcolor{blue!70!black}{$\mathbf{\underline{.712}}$$\uparrow$} & 
\textcolor{blue!70!black}{$.451$$\uparrow$} \\
\bottomrule
\end{tabular}%
}
\caption{SEAHORSE Pearson $r$ for dimension \textbf{Comprehensible} per language, using English prompts. Baseline uses $\sigma{=}0$ (no 
steering); steered uses the best $\sigma$ selected on the validation split; avgvec zero-shot uses a fixed vector averaged across all 
source languages. \textbf{Bold} indicates the highest value per language across the entire table; \underline{underline} indicates the 
highest value per language within each model. \textcolor{blue!70!black}{Blue}/$\uparrow$ (\textcolor{red!70!black}{red}/$\downarrow$) on 
steered rows indicates improvement (degradation) relative to the baseline.}
\label{tab:seahorse_perdim_question1_en}
\end{table*}

\begin{table*}[t]
\centering
\resizebox{\textwidth}{!}{%
\begin{tabular}{ll|c|c|c|c|c}
\textbf{} & \textbf{} & \textbf{de} & \textbf{es} & \textbf{ru} & \textbf{tr} & \textbf{vi} \\
\toprule
\multicolumn{7}{l}{\textbf{GPTScore}} \\
\midrule
\multirow{3}{*}{\texttt{Llama-3.2-3B}} & Baseline (EN-prompt) & $-.147$ & $.068$ & $.178$ & $\underline{.095}$ & $.079$ \\
 & + vector steer (EN-prompt) & \textcolor{blue!70!black}{$-.145$$\uparrow$} & \textcolor{red!70!black}{$.067$$\downarrow$} & 
\textcolor{blue!70!black}{$.185$$\uparrow$} & \textcolor{red!70!black}{$.094$$\downarrow$} & 
\textcolor{blue!70!black}{$\underline{.079}$$\uparrow$} \\
 & + avgvec zero-shot (EN-prompt) & \textcolor{blue!70!black}{$\underline{-.139}$$\uparrow$} & 
\textcolor{blue!70!black}{$\underline{.075}$$\uparrow$} & \textcolor{blue!70!black}{$\underline{.191}$$\uparrow$} & 
\textcolor{red!70!black}{$\underline{.095}$$\downarrow$} & \textcolor{red!70!black}{$.075$$\downarrow$} \\
\midrule
\multirow{3}{*}{\texttt{aya-expanse-8b}} & Baseline (EN-prompt) & $-.207$ & $.033$ & $.234$ & $.168$ & $.193$ \\
 & + vector steer (EN-prompt) & \textcolor{blue!70!black}{$-.206$$\uparrow$} & \textcolor{blue!70!black}{$.038$$\uparrow$} & 
\textcolor{blue!70!black}{$.242$$\uparrow$} & \textcolor{blue!70!black}{$.176$$\uparrow$} & \textcolor{blue!70!black}{$.196$$\uparrow$} 
\\
 & + avgvec zero-shot (EN-prompt) & \textcolor{blue!70!black}{$\underline{-.199}$$\uparrow$} & 
\textcolor{blue!70!black}{$\underline{.040}$$\uparrow$} & \textcolor{blue!70!black}{$\underline{.245}$$\uparrow$} & 
\textcolor{blue!70!black}{$\underline{.177}$$\uparrow$} & \textcolor{blue!70!black}{$\underline{.200}$$\uparrow$} \\
\midrule
\multicolumn{7}{l}{\textbf{Direct Prompting}} \\
\midrule
\multirow{3}{*}{\texttt{Llama-3.2-3B}} & Baseline (EN-prompt) & $\underline{.088}$ & $.012$ & $.057$ & $\underline{.045}$ & $.059$ \\
 & + vector steer (EN-prompt) & \textcolor{red!70!black}{$.081$$\downarrow$} & \textcolor{blue!70!black}{$\underline{.017}$$\uparrow$} & 
\textcolor{blue!70!black}{$\underline{.060}$$\uparrow$} & \textcolor{red!70!black}{$.043$$\downarrow$} & 
\textcolor{blue!70!black}{$\underline{.064}$$\uparrow$} \\
 & + avgvec zero-shot (EN-prompt) & --- & --- & --- & --- & --- \\
\midrule
\multirow{3}{*}{\texttt{aya-expanse-8b}} & Baseline (EN-prompt) & $\underline{.210}$ & $\underline{.147}$ & $.173$ & $.179$ & 
$\underline{.180}$ \\
 & + vector steer (EN-prompt) & \textcolor{red!70!black}{$.206$$\downarrow$} & \textcolor{red!70!black}{$\underline{.146}$$\downarrow$} & 
\textcolor{blue!70!black}{$\underline{.189}$$\uparrow$} & \textcolor{blue!70!black}{$\underline{.195}$$\uparrow$} & 
\textcolor{red!70!black}{$.177$$\downarrow$} \\
 & + avgvec zero-shot (EN-prompt) & --- & --- & --- & --- & --- \\
\midrule
\multicolumn{7}{l}{\textbf{COMET}} \\
\midrule
\multirow{3}{*}{\texttt{wmt22-comet-da}} & Baseline & $.301$ & $.338$ & $.406$ & $.414$ & $.427$ \\
 & + vector steer & \textcolor{blue!70!black}{$\mathbf{\underline{.347}}$$\uparrow$} & 
\textcolor{blue!70!black}{$\mathbf{\underline{.513}}$$\uparrow$} & \textcolor{blue!70!black}{$.468$$\uparrow$} & 
\textcolor{blue!70!black}{$.450$$\uparrow$} & \textcolor{blue!70!black}{$\mathbf{\underline{.493}}$$\uparrow$} \\
 & + avgvec zero-shot & \textcolor{blue!70!black}{$.345$$\uparrow$} & \textcolor{blue!70!black}{$.428$$\uparrow$} & 
\textcolor{blue!70!black}{$\mathbf{\underline{.470}}$$\uparrow$} & \textcolor{blue!70!black}{$\mathbf{\underline{.468}}$$\uparrow$} & 
\textcolor{blue!70!black}{$.488$$\uparrow$} \\
\bottomrule
\end{tabular}%
}
\caption{SEAHORSE Pearson $r$ for dimension \textbf{Non-repetitive} per language, using English prompts. Baseline uses $\sigma{=}0$ (no 
steering); steered uses the best $\sigma$ selected on the validation split; avgvec zero-shot uses a fixed vector averaged across all 
source languages. \textbf{Bold} indicates the highest value per language across the entire table; \underline{underline} indicates the 
highest value per language within each model. \textcolor{blue!70!black}{Blue}/$\uparrow$ (\textcolor{red!70!black}{red}/$\downarrow$) on 
steered rows indicates improvement (degradation) relative to the baseline.}
\label{tab:seahorse_perdim_question2_en}
\end{table*}

\begin{table*}[t]
\centering
\resizebox{\textwidth}{!}{%
\begin{tabular}{ll|c|c|c|c|c}
\textbf{} & \textbf{} & \textbf{de} & \textbf{es} & \textbf{ru} & \textbf{tr} & \textbf{vi} \\
\toprule
\multicolumn{7}{l}{\textbf{GPTScore}} \\
\midrule
\multirow{3}{*}{\texttt{Llama-3.2-3B}} & Baseline (EN-prompt) & $-.073$ & $\underline{-.023}$ & $.190$ & $.090$ & $.124$ \\
 & + vector steer (EN-prompt) & \textcolor{red!70!black}{$-.074$$\downarrow$} & \textcolor{blue!70!black}{$\underline{-.023}$$\uparrow$} 
& \textcolor{blue!70!black}{$.197$$\uparrow$} & \textcolor{red!70!black}{$.090$$\downarrow$} & 
\textcolor{blue!70!black}{$\underline{.125}$$\uparrow$} \\
 & + avgvec zero-shot (EN-prompt) & \textcolor{blue!70!black}{$\underline{-.070}$$\uparrow$} & 
\textcolor{red!70!black}{$-.033$$\downarrow$} & \textcolor{blue!70!black}{$\mathbf{\underline{.198}}$$\uparrow$} & 
\textcolor{blue!70!black}{$\underline{.092}$$\uparrow$} & \textcolor{red!70!black}{$.121$$\downarrow$} \\
\midrule
\multirow{3}{*}{\texttt{aya-expanse-8b}} & Baseline (EN-prompt) & $-.101$ & $.065$ & $.140$ & $.121$ & $.171$ \\
 & + vector steer (EN-prompt) & \textcolor{blue!70!black}{$-.101$$\uparrow$} & 
\textcolor{blue!70!black}{$\mathbf{\underline{.068}}$$\uparrow$} & \textcolor{blue!70!black}{$.148$$\uparrow$} & 
\textcolor{blue!70!black}{$.126$$\uparrow$} & \textcolor{blue!70!black}{$.174$$\uparrow$} \\
 & + avgvec zero-shot (EN-prompt) & \textcolor{blue!70!black}{$\underline{-.096}$$\uparrow$} & 
\textcolor{blue!70!black}{$.066$$\uparrow$} & \textcolor{blue!70!black}{$\underline{.150}$$\uparrow$} & 
\textcolor{blue!70!black}{$\underline{.128}$$\uparrow$} & \textcolor{blue!70!black}{$\underline{.177}$$\uparrow$} \\
\midrule
\multicolumn{7}{l}{\textbf{Direct Prompting}} \\
\midrule
\multirow{3}{*}{\texttt{Llama-3.2-3B}} & Baseline (EN-prompt) & $.013$ & $.010$ & $\underline{.010}$ & $.023$ & $-.038$ \\
 & + vector steer (EN-prompt) & \textcolor{blue!70!black}{$\underline{.014}$$\uparrow$} & 
\textcolor{blue!70!black}{$\underline{.016}$$\uparrow$} & \textcolor{red!70!black}{$.009$$\downarrow$} & 
\textcolor{blue!70!black}{$\underline{.025}$$\uparrow$} & \textcolor{blue!70!black}{$\underline{-.031}$$\uparrow$} \\
 & + avgvec zero-shot (EN-prompt) & --- & --- & --- & --- & --- \\
\midrule
\multirow{3}{*}{\texttt{aya-expanse-8b}} & Baseline (EN-prompt) & $\mathbf{\underline{.159}}$ & $-.048$ & $\underline{.139}$ & 
$\underline{.152}$ & $\underline{.129}$ \\
 & + vector steer (EN-prompt) & \textcolor{red!70!black}{$\mathbf{\underline{.159}}$$\downarrow$} & 
\textcolor{blue!70!black}{$\underline{-.048}$$\uparrow$} & \textcolor{red!70!black}{$\underline{.139}$$\downarrow$} & 
\textcolor{blue!70!black}{$\underline{.153}$$\uparrow$} & \textcolor{red!70!black}{$.126$$\downarrow$} \\
 & + avgvec zero-shot (EN-prompt) & --- & --- & --- & --- & --- \\
\midrule
\multicolumn{7}{l}{\textbf{COMET}} \\
\midrule
\multirow{3}{*}{\texttt{wmt22-comet-da}} & Baseline & $.090$ & $\underline{.045}$ & $\underline{.197}$ & $.345$ & $.245$ \\
 & + vector steer & \textcolor{blue!70!black}{$\underline{.128}$$\uparrow$} & \textcolor{red!70!black}{$-.001$$\downarrow$} & 
\textcolor{red!70!black}{$.170$$\downarrow$} & \textcolor{red!70!black}{$.308$$\downarrow$} & 
\textcolor{red!70!black}{$.237$$\downarrow$} \\
 & + avgvec zero-shot & \textcolor{blue!70!black}{$.119$$\uparrow$} & \textcolor{red!70!black}{$.041$$\downarrow$} & 
\textcolor{red!70!black}{$.183$$\downarrow$} & \textcolor{blue!70!black}{$\mathbf{\underline{.352}}$$\uparrow$} & 
\textcolor{blue!70!black}{$\mathbf{\underline{.246}}$$\uparrow$} \\
\bottomrule
\end{tabular}%
}
\caption{SEAHORSE Pearson $r$ for dimension \textbf{Grammatical} per language, using English prompts. Baseline uses $\sigma{=}0$ (no 
steering); steered uses the best $\sigma$ selected on the validation split; avgvec zero-shot uses a fixed vector averaged across all 
source languages. \textbf{Bold} indicates the highest value per language across the entire table; \underline{underline} indicates the 
highest value per language within each model. \textcolor{blue!70!black}{Blue}/$\uparrow$ (\textcolor{red!70!black}{red}/$\downarrow$) on 
steered rows indicates improvement (degradation) relative to the baseline.}
\label{tab:seahorse_perdim_question3_en}
\end{table*}

\begin{table*}[t]
\centering
\resizebox{\textwidth}{!}{%
\begin{tabular}{ll|c|c|c|c|c}
\textbf{} & \textbf{} & \textbf{de} & \textbf{es} & \textbf{ru} & \textbf{tr} & \textbf{vi} \\
\toprule
\multicolumn{7}{l}{\textbf{GPTScore}} \\
\midrule
\multirow{3}{*}{\texttt{Llama-3.2-3B}} & Baseline (EN-prompt) & $\underline{.279}$ & $.368$ & $.334$ & $.336$ & $.278$ \\
 & + vector steer (EN-prompt) & \textcolor{red!70!black}{$.268$$\downarrow$} & \textcolor{blue!70!black}{$.371$$\uparrow$} & 
\textcolor{blue!70!black}{$\mathbf{\underline{.342}}$$\uparrow$} & \textcolor{red!70!black}{$.334$$\downarrow$} & 
\textcolor{blue!70!black}{$\underline{.284}$$\uparrow$} \\
 & + avgvec zero-shot (EN-prompt) & \textcolor{red!70!black}{$.272$$\downarrow$} & 
\textcolor{blue!70!black}{$\mathbf{\underline{.388}}$$\uparrow$} & \textcolor{blue!70!black}{$.341$$\uparrow$} & 
\textcolor{blue!70!black}{$\mathbf{\underline{.341}}$$\uparrow$} & \textcolor{blue!70!black}{$.280$$\uparrow$} \\
\midrule
\multirow{3}{*}{\texttt{aya-expanse-8b}} & Baseline (EN-prompt) & $.249$ & $.101$ & $.190$ & $.197$ & $.182$ \\
 & + vector steer (EN-prompt) & \textcolor{blue!70!black}{$.254$$\uparrow$} & \textcolor{blue!70!black}{$.107$$\uparrow$} & 
\textcolor{blue!70!black}{$.196$$\uparrow$} & \textcolor{blue!70!black}{$\underline{.208}$$\uparrow$} & 
\textcolor{blue!70!black}{$.186$$\uparrow$} \\
 & + avgvec zero-shot (EN-prompt) & \textcolor{blue!70!black}{$\underline{.258}$$\uparrow$} & 
\textcolor{blue!70!black}{$\underline{.111}$$\uparrow$} & \textcolor{blue!70!black}{$\underline{.198}$$\uparrow$} & 
\textcolor{blue!70!black}{$\underline{.208}$$\uparrow$} & \textcolor{blue!70!black}{$\underline{.189}$$\uparrow$} \\
\midrule
\multicolumn{7}{l}{\textbf{Direct Prompting}} \\
\midrule
\multirow{3}{*}{\texttt{Llama-3.2-3B}} & Baseline (EN-prompt) & $\underline{-.040}$ & $-.099$ & $\underline{-.044}$ & $\underline{-.062}$ 
& $-.051$ \\
 & + vector steer (EN-prompt) & \textcolor{blue!70!black}{$\underline{-.039}$$\uparrow$} & 
\textcolor{blue!70!black}{$\underline{-.094}$$\uparrow$} & \textcolor{red!70!black}{$-.045$$\downarrow$} & 
\textcolor{red!70!black}{$-.066$$\downarrow$} & \textcolor{blue!70!black}{$\underline{-.048}$$\uparrow$} \\
 & + avgvec zero-shot (EN-prompt) & --- & --- & --- & --- & --- \\
\midrule
\multirow{3}{*}{\texttt{aya-expanse-8b}} & Baseline (EN-prompt) & $.293$ & $.344$ & $\underline{.266}$ & $\underline{.222}$ & 
$\mathbf{\underline{.314}}$ \\
 & + vector steer (EN-prompt) & \textcolor{blue!70!black}{$\mathbf{\underline{.295}}$$\uparrow$} & 
\textcolor{blue!70!black}{$\underline{.345}$$\uparrow$} & \textcolor{red!70!black}{$.265$$\downarrow$} & 
\textcolor{red!70!black}{$.219$$\downarrow$} & \textcolor{blue!70!black}{$\mathbf{\underline{.315}}$$\uparrow$} \\
 & + avgvec zero-shot (EN-prompt) & --- & --- & --- & --- & --- \\
\midrule
\multicolumn{7}{l}{\textbf{COMET}} \\
\midrule
\multirow{3}{*}{\texttt{wmt22-comet-da}} & Baseline & $.168$ & $.219$ & $\underline{.215}$ & $\underline{.204}$ & $.112$ \\
 & + vector steer & \textcolor{blue!70!black}{$\underline{.226}$$\uparrow$} & \textcolor{blue!70!black}{$\underline{.258}$$\uparrow$} & 
\textcolor{red!70!black}{$.202$$\downarrow$} & \textcolor{red!70!black}{$.176$$\downarrow$} & 
\textcolor{blue!70!black}{$\underline{.123}$$\uparrow$} \\
 & + avgvec zero-shot & \textcolor{blue!70!black}{$.209$$\uparrow$} & \textcolor{blue!70!black}{$.250$$\uparrow$} & 
\textcolor{red!70!black}{$.211$$\downarrow$} & \textcolor{red!70!black}{$.195$$\downarrow$} & \textcolor{blue!70!black}{$.121$$\uparrow$} 
\\
\bottomrule
\end{tabular}%
}
\caption{SEAHORSE Pearson $r$ for dimension \textbf{Attributable} per language, using English prompts. Baseline uses $\sigma{=}0$ (no 
steering); steered uses the best $\sigma$ selected on the validation split; avgvec zero-shot uses a fixed vector averaged across all 
source languages. \textbf{Bold} indicates the highest value per language across the entire table; \underline{underline} indicates the 
highest value per language within each model. \textcolor{blue!70!black}{Blue}/$\uparrow$ (\textcolor{red!70!black}{red}/$\downarrow$) on 
steered rows indicates improvement (degradation) relative to the baseline.}
\label{tab:seahorse_perdim_question4_en}
\end{table*}

\begin{table*}[t]
\centering
\resizebox{\textwidth}{!}{%
\begin{tabular}{ll|c|c|c|c|c}
\textbf{} & \textbf{} & \textbf{de} & \textbf{es} & \textbf{ru} & \textbf{tr} & \textbf{vi} \\
\toprule
\multicolumn{7}{l}{\textbf{GPTScore}} \\
\midrule
\multirow{3}{*}{\texttt{Llama-3.2-3B}} & Baseline (EN-prompt) & $\underline{.023}$ & $.267$ & $.303$ & $\underline{.290}$ & 
$\underline{.029}$ \\
 & + vector steer (EN-prompt) & \textcolor{red!70!black}{$.021$$\downarrow$} & \textcolor{blue!70!black}{$.271$$\uparrow$} & 
\textcolor{blue!70!black}{$\underline{.308}$$\uparrow$} & \textcolor{red!70!black}{$.285$$\downarrow$} & 
\textcolor{red!70!black}{$.021$$\downarrow$} \\
 & + avgvec zero-shot (EN-prompt) & \textcolor{red!70!black}{$.020$$\downarrow$} & 
\textcolor{blue!70!black}{$\underline{.272}$$\uparrow$} & \textcolor{blue!70!black}{$\underline{.308}$$\uparrow$} & 
\textcolor{red!70!black}{$.288$$\downarrow$} & \textcolor{red!70!black}{$.022$$\downarrow$} \\
\midrule
\multirow{3}{*}{\texttt{aya-expanse-8b}} & Baseline (EN-prompt) & $.073$ & $.258$ & $.284$ & $.317$ & $.136$ \\
 & + vector steer (EN-prompt) & \textcolor{blue!70!black}{$.074$$\uparrow$} & \textcolor{blue!70!black}{$.261$$\uparrow$} & 
\textcolor{blue!70!black}{$\underline{.289}$$\uparrow$} & \textcolor{blue!70!black}{$.324$$\uparrow$} & 
\textcolor{blue!70!black}{$.138$$\uparrow$} \\
 & + avgvec zero-shot (EN-prompt) & \textcolor{blue!70!black}{$\underline{.075}$$\uparrow$} & 
\textcolor{blue!70!black}{$\underline{.263}$$\uparrow$} & \textcolor{blue!70!black}{$\underline{.289}$$\uparrow$} & 
\textcolor{blue!70!black}{$\underline{.325}$$\uparrow$} & \textcolor{blue!70!black}{$\underline{.140}$$\uparrow$} \\
\midrule
\multicolumn{7}{l}{\textbf{Direct Prompting}} \\
\midrule
\multirow{3}{*}{\texttt{Llama-3.2-3B}} & Baseline (EN-prompt) & $\underline{.052}$ & $.013$ & $-.030$ & $-.039$ & $-.013$ \\
 & + vector steer (EN-prompt) & \textcolor{red!70!black}{$\underline{.052}$$\downarrow$} & 
\textcolor{blue!70!black}{$\underline{.025}$$\uparrow$} & \textcolor{blue!70!black}{$\underline{-.024}$$\uparrow$} & 
\textcolor{blue!70!black}{$\underline{-.035}$$\uparrow$} & \textcolor{blue!70!black}{$\underline{-.006}$$\uparrow$} \\
 & + avgvec zero-shot (EN-prompt) & --- & --- & --- & --- & --- \\
\midrule
\multirow{3}{*}{\texttt{aya-expanse-8b}} & Baseline (EN-prompt) & $.182$ & $\underline{.303}$ & $\underline{.274}$ & $.241$ & 
$\underline{.304}$ \\
 & + vector steer (EN-prompt) & \textcolor{blue!70!black}{$\underline{.182}$$\uparrow$} & 
\textcolor{red!70!black}{$\underline{.302}$$\downarrow$} & \textcolor{red!70!black}{$.272$$\downarrow$} & 
\textcolor{blue!70!black}{$\underline{.242}$$\uparrow$} & \textcolor{red!70!black}{$\underline{.303}$$\downarrow$} \\
 & + avgvec zero-shot (EN-prompt) & --- & --- & --- & --- & --- \\
\midrule
\multicolumn{7}{l}{\textbf{COMET}} \\
\midrule
\multirow{3}{*}{\texttt{wmt22-comet-da}} & Baseline & $.324$ & $.340$ & $\mathbf{\underline{.419}}$ & $\mathbf{\underline{.378}}$ & 
$\mathbf{\underline{.331}}$ \\
 & + vector steer & \textcolor{red!70!black}{$.320$$\downarrow$} & \textcolor{red!70!black}{$.331$$\downarrow$} & 
\textcolor{red!70!black}{$.346$$\downarrow$} & \textcolor{red!70!black}{$.335$$\downarrow$} & 
\textcolor{red!70!black}{$.287$$\downarrow$} \\
 & + avgvec zero-shot & \textcolor{blue!70!black}{$\mathbf{\underline{.325}}$$\uparrow$} & 
\textcolor{blue!70!black}{$\mathbf{\underline{.357}}$$\uparrow$} & \textcolor{red!70!black}{$.376$$\downarrow$} & 
\textcolor{red!70!black}{$.346$$\downarrow$} & \textcolor{red!70!black}{$.293$$\downarrow$} \\
\bottomrule
\end{tabular}%
}
\caption{SEAHORSE Pearson $r$ for dimension \textbf{Main ideas} per language, using English prompts. Baseline uses $\sigma{=}0$ (no 
steering); steered uses the best $\sigma$ selected on the validation split; avgvec zero-shot uses a fixed vector averaged across all 
source languages. \textbf{Bold} indicates the highest value per language across the entire table; \underline{underline} indicates the 
highest value per language within each model. \textcolor{blue!70!black}{Blue}/$\uparrow$ (\textcolor{red!70!black}{red}/$\downarrow$) on 
steered rows indicates improvement (degradation) relative to the baseline.}
\label{tab:seahorse_perdim_question5_en}
\end{table*}

\begin{table*}[t]
\centering
\resizebox{\textwidth}{!}{%
\begin{tabular}{ll|c|c|c|c|c}
\textbf{} & \textbf{} & \textbf{de} & \textbf{es} & \textbf{ru} & \textbf{tr} & \textbf{vi} \\
\toprule
\multicolumn{7}{l}{\textbf{GPTScore}} \\
\midrule
\multirow{3}{*}{\texttt{Llama-3.2-3B}} & Baseline (EN-prompt) & $\underline{.045}$ & $.279$ & $.303$ & $.303$ & $\underline{.079}$ \\
 & + vector steer (EN-prompt) & \textcolor{red!70!black}{$.035$$\downarrow$} & \textcolor{blue!70!black}{$.284$$\uparrow$} & 
\textcolor{blue!70!black}{$.312$$\uparrow$} & \textcolor{red!70!black}{$.301$$\downarrow$} & \textcolor{red!70!black}{$.075$$\downarrow$} 
\\
 & + avgvec zero-shot (EN-prompt) & \textcolor{red!70!black}{$.037$$\downarrow$} & 
\textcolor{blue!70!black}{$\mathbf{\underline{.290}}$$\uparrow$} & \textcolor{blue!70!black}{$\underline{.314}$$\uparrow$} & 
\textcolor{blue!70!black}{$\mathbf{\underline{.306}}$$\uparrow$} & \textcolor{red!70!black}{$.075$$\downarrow$} \\
\midrule
\multirow{3}{*}{\texttt{aya-expanse-8b}} & Baseline (EN-prompt) & $.058$ & $.195$ & $.237$ & $.223$ & $.102$ \\
 & + vector steer (EN-prompt) & \textcolor{blue!70!black}{$\underline{.062}$$\uparrow$} & \textcolor{blue!70!black}{$.199$$\uparrow$} & 
\textcolor{blue!70!black}{$.244$$\uparrow$} & \textcolor{blue!70!black}{$.235$$\uparrow$} & \textcolor{blue!70!black}{$.106$$\uparrow$} 
\\
 & + avgvec zero-shot (EN-prompt) & \textcolor{blue!70!black}{$\underline{.062}$$\uparrow$} & 
\textcolor{blue!70!black}{$\underline{.202}$$\uparrow$} & \textcolor{blue!70!black}{$\underline{.246}$$\uparrow$} & 
\textcolor{blue!70!black}{$\underline{.236}$$\uparrow$} & \textcolor{blue!70!black}{$\underline{.109}$$\uparrow$} \\
\midrule
\multicolumn{7}{l}{\textbf{Direct Prompting}} \\
\midrule
\multirow{3}{*}{\texttt{Llama-3.2-3B}} & Baseline (EN-prompt) & $\underline{.094}$ & $\underline{.042}$ & $\underline{.034}$ & 
$\underline{.018}$ & $-.008$ \\
 & + vector steer (EN-prompt) & \textcolor{red!70!black}{$.092$$\downarrow$} & \textcolor{red!70!black}{$.036$$\downarrow$} & 
\textcolor{red!70!black}{$.029$$\downarrow$} & \textcolor{red!70!black}{$.016$$\downarrow$} & 
\textcolor{blue!70!black}{$\underline{-.001}$$\uparrow$} \\
 & + avgvec zero-shot (EN-prompt) & --- & --- & --- & --- & --- \\
\midrule
\multirow{3}{*}{\texttt{aya-expanse-8b}} & Baseline (EN-prompt) & $.197$ & $\underline{.274}$ & $\underline{.257}$ & $\underline{.153}$ & 
$.220$ \\
 & + vector steer (EN-prompt) & \textcolor{blue!70!black}{$\underline{.201}$$\uparrow$} & 
\textcolor{blue!70!black}{$\underline{.274}$$\uparrow$} & \textcolor{blue!70!black}{$\underline{.257}$$\uparrow$} & 
\textcolor{red!70!black}{$.146$$\downarrow$} & \textcolor{blue!70!black}{$\mathbf{\underline{.224}}$$\uparrow$} \\
 & + avgvec zero-shot (EN-prompt) & --- & --- & --- & --- & --- \\
\midrule
\multicolumn{7}{l}{\textbf{COMET}} \\
\midrule
\multirow{3}{*}{\texttt{wmt22-comet-da}} & Baseline & $\mathbf{\underline{.307}}$ & $.233$ & $\mathbf{\underline{.334}}$ & 
$\underline{.288}$ & $\underline{.201}$ \\
 & + vector steer & \textcolor{red!70!black}{$.292$$\downarrow$} & \textcolor{blue!70!black}{$.253$$\uparrow$} & 
\textcolor{red!70!black}{$.296$$\downarrow$} & \textcolor{red!70!black}{$.267$$\downarrow$} & 
\textcolor{red!70!black}{$.179$$\downarrow$} \\
 & + avgvec zero-shot & \textcolor{red!70!black}{$.306$$\downarrow$} & \textcolor{blue!70!black}{$\underline{.254}$$\uparrow$} & 
\textcolor{red!70!black}{$.319$$\downarrow$} & \textcolor{red!70!black}{$.277$$\downarrow$} & 
\textcolor{red!70!black}{$.182$$\downarrow$} \\
\bottomrule
\end{tabular}%
}
\caption{SEAHORSE Pearson $r$ for dimension \textbf{Concise} per language, using English prompts. Baseline uses $\sigma{=}0$ (no 
steering); steered uses the best $\sigma$ selected on the validation split; avgvec zero-shot uses a fixed vector averaged across all 
source languages. \textbf{Bold} indicates the highest value per language across the entire table; \underline{underline} indicates the 
highest value per language within each model. \textcolor{blue!70!black}{Blue}/$\uparrow$ (\textcolor{red!70!black}{red}/$\downarrow$) on 
steered rows indicates improvement (degradation) relative to the baseline.}
\label{tab:seahorse_perdim_question6_en}
\end{table*}

\clearpage

\section{Optimal Hyperparameters}
\begin{table*}[t]
\centering
\resizebox{\textwidth}{!}{%
\begin{tabular}{ll|cc|cc|cc|cc|cc}
 & & \multicolumn{2}{c}{\textbf{de}} & \multicolumn{2}{c}{\textbf{es}} & \multicolumn{2}{c}{\textbf{ru}} & 
\multicolumn{2}{c}{\textbf{tr}} & \multicolumn{2}{c}{\textbf{vi}} \\
 & & layer & $\sigma$ & layer & $\sigma$ & layer & $\sigma$ & layer & $\sigma$ & layer & $\sigma$ \\
\toprule
\multicolumn{12}{l}{\textbf{GPTScore}} \\
\midrule
\multirow{2}{*}{\texttt{Llama-3.2-3B}} & EN-prompt & 27 & $-2.0$ & 27 & $+2.0$ & 27 & $-1.0$ & 27 & $-5.0$ & 22 & $-2.0$ \\
 & src-prompt & 27 & $-2.0$ & 27 & $+1.0$ & 27 & $-1.0$ & 27 & $-3.0$ & 27 & $+1.0$ \\
\midrule
\multirow{2}{*}{\texttt{aya-expanse-8b}} & EN-prompt & 27 & $-5.0$ & 31 & $-3.0$ & 31 & $-2.0$ & 31 & $-3.0$ & 31 & $-2.0$ \\
 & src-prompt & 31 & $-3.0$ & 31 & $-2.0$ & 31 & $-1.0$ & 31 & $-2.0$ & 31 & $-1.0$ \\
\midrule
\multicolumn{12}{l}{\textbf{Direct Prompting}} \\
\midrule
\multirow{2}{*}{\texttt{Llama-3.2-3B}} & EN-prompt & 27 & $+5.0$ & 27 & $-5.0$ & 24 & $+5.0$ & 25 & $+5.0$ & 22 & $+3.0$ \\
 & src-prompt & 25 & $+4.0$ & 27 & $-5.0$ & 21 & $+2.0$ & 23 & $+4.0$ & 26 & $-4.0$ \\
\midrule
\multirow{2}{*}{\texttt{aya-expanse-8b}} & EN-prompt & 29 & $-5.0$ & 31 & $+5.0$ & 26 & $-5.0$ & 26 & $-5.0$ & 25 & $-5.0$ \\
 & src-prompt & 26 & $-5.0$ & 26 & $-5.0$ & 26 & $-2.0$ & 29 & $-3.0$ & 25 & $-5.0$ \\
\bottomrule
\end{tabular}%
}
\caption{Best intervention layer and steering $\sigma$ for LLM-based SEAHORSE evaluation. Layer and $\sigma$ are selected jointly on the 
validation split by mean Pearson~$r$ across six quality dimensions. Avgvec rows use a fixed $\sigma$ and the layer that maximised 
cross-lingual vector similarity.}
\label{tab:seahorse_hyperparams_llm}
\end{table*}

\begin{table}[t]
\centering
\begin{tabular}{l|c|c|c|c|c}
& \textbf{de} & \textbf{es} & \textbf{ru} & \textbf{tr} & \textbf{vi} \\
\toprule
\multicolumn{6}{l}{\textbf{COMET (wmt22-comet-da)}} \\
\midrule
vector steer & $-5.0$ & $-5.0$ & $-2.0$ & $+5.0$ & $-2.0$ \\
\bottomrule
\end{tabular}
\caption{Best steering $\sigma$ for COMET (\texttt{wmt22-comet-da}) SEAHORSE evaluation. COMET steers the pooling-space embedding 
directly (no per-layer selection); $\sigma$ is chosen on the validation split by mean Pearson~$r$ across six quality dimensions.}
\label{tab:seahorse_hyperparams_comet}
\end{table}

\clearpage
\section{Prompts}

\begin{figure*}[t]
  \centering
    \includegraphics[width=\textwidth]{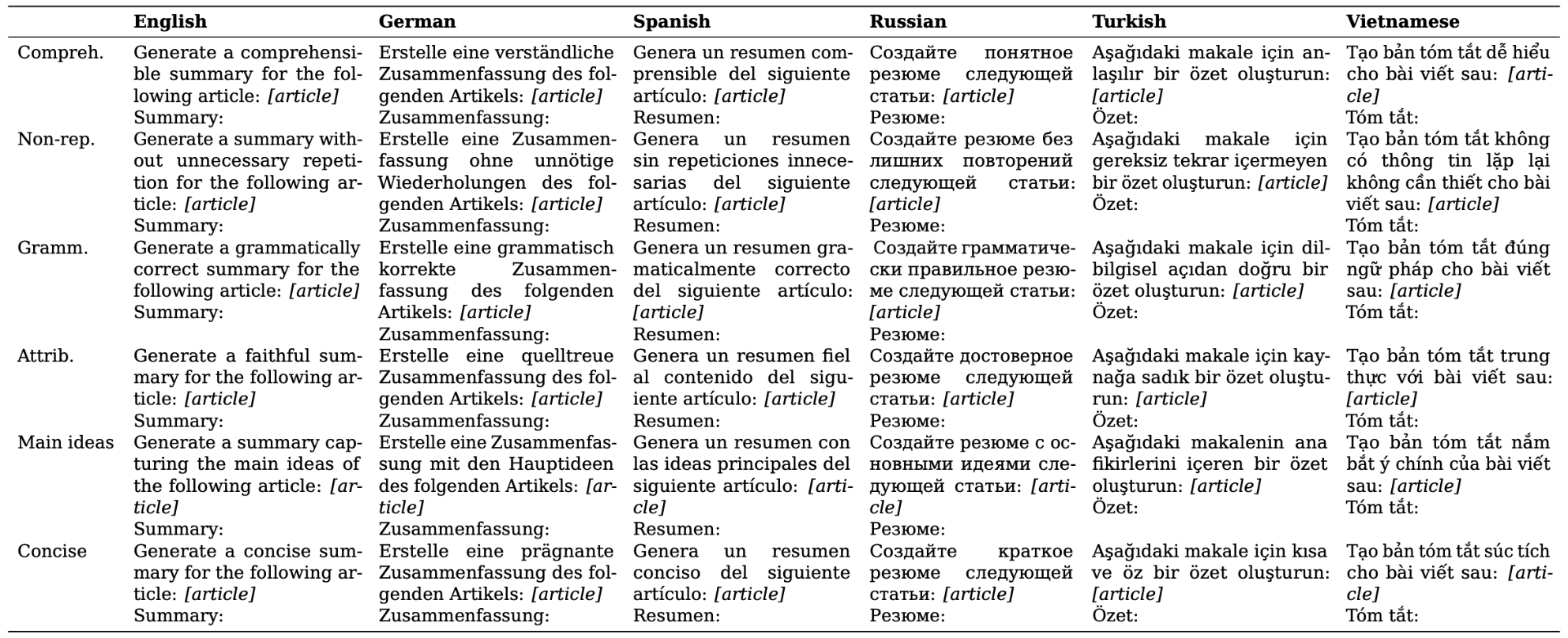}
    \caption{GPTScore prompt templates per quality dimension and language.
  \textit{[article]} is replaced by the source article; the second line (after the newline) is the continuation cue used for scoring.}
  \label{tab:gptscore_prompts}
\end{figure*}

\begin{figure*}[t]
  \centering
    \includegraphics[width=\textwidth]{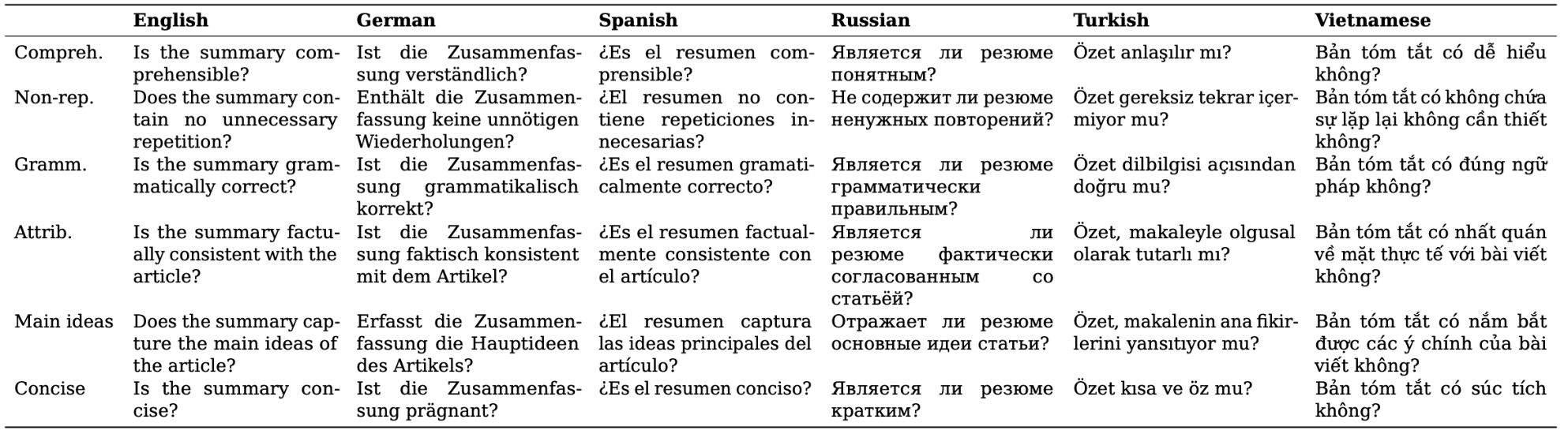}
  \caption{Direct Prompting (G-eval) question per quality dimension and language.}
  \label{tab:geval_questions}
\end{figure*}

\begin{figure*}[t]
  \centering
    \includegraphics[width=\textwidth]{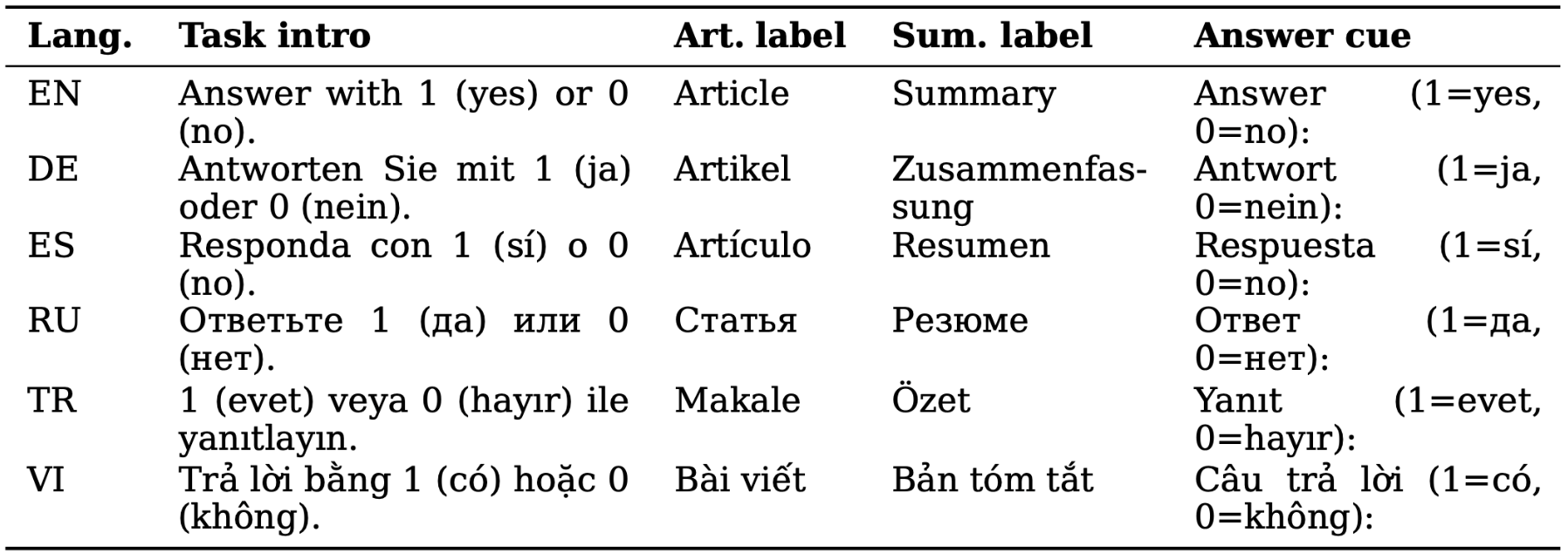}
    \caption{Fixed sentence-level labels used in Direct Prompting per language.}
  \label{tab:geval_labels}
\end{figure*}

\clearpage

\section{Dataset}
\label{app:data}

Data is released under a  CC-BY 4.0 license, with which we comply fully. 

\begin{table}[h]
\centering
\resizebox{\columnwidth}{!}{%
\begin{tabular}{lrrrrrr}
\toprule
System & DE & EN & ES & RU & TR & VI \\
\midrule
T5-Base & 0 & 236 & 0 & 0 & 0 & 0 \\
T5-Base (250s) & 0 & 241 & 0 & 0 & 0 & 0 \\
T5-XXL & 0 & 247 & 0 & 0 & 0 & 0 \\
mT5-Small & 231 & 239 & 234 & 222 & 218 & 234 \\
mT5-Small (250s) & 226 & 243 & 239 & 218 & 209 & 229 \\
mT5-XXL & 228 & 239 & 242 & 208 & 208 & 238 \\
PaLM (1-shot) & 241 & 250 & 239 & 241 & 207 & 223 \\
PaLM (fine-tuned) & 219 & 249 & 245 & 241 & 180 & 230 \\
Reference & 228 & 239 & 240 & 213 & 210 & 238 \\
\midrule
Total outputs & 1,373 & 2,183 & 1,439 & 1,343 & 1,232 & 1,392 \\
\bottomrule
\end{tabular}%
}
\caption{Number of outputs samples, annotated with human judgments for the Seahorse validation set.}
\label{tab:seahorse-validation}
\end{table}

\begin{table}[h!]
\centering
\resizebox{\columnwidth}{!}{%
\begin{tabular}{lrrrrrr}
\toprule
System & DE & EN & ES & RU & TR & VI \\
\midrule
T5-Base & 0 & 484 & 0 & 0 & 0 & 0 \\
T5-Base (250s) & 0 & 482 & 0 & 0 & 0 & 0 \\
T5-XXL & 0 & 476 & 0 & 0 & 0 & 0 \\
mT5-Small & 451 & 482 & 474 & 449 & 495 & 497 \\
mT5-Small (250s) & 459 & 484 & 477 & 427 & 488 & 490 \\
mT5-XXL & 443 & 481 & 479 & 450 & 485 & 492 \\
PaLM (1-shot) & 463 & 495 & 486 & 486 & 440 & 438 \\
PaLM (fine-tuned) & 434 & 487 & 473 & 485 & 409 & 483 \\
Reference & 435 & 484 & 465 & 439 & 486 & 494 \\
\midrule
Total outputs & 2,685 & 4,355 & 2,854 & 2,736 & 2,803 & 2,894 \\
\bottomrule
\end{tabular}%
}
\caption{Number of outputs samples, annotated with human judgments for the Seahorse test set.}
\label{tab:seahorse-test}
\end{table}

\clearpage

\section{Source prompting results}
\label{app:src_results}
Table~\ref{tab:seahorse_condensed_src} presents the results when using a machine-translated prompt in the source language.

\begin{table*}[t]
\centering
\resizebox{\textwidth}{!}{%
\begin{tabular}{ll|c|c|c|c|c}
\textbf{} & \textbf{} & \textbf{de} & \textbf{es} & \textbf{ru} & \textbf{tr} & \textbf{vi} \\
\toprule
\multicolumn{7}{l}{\textbf{GPTScore}} \\
\midrule
\multirow{2}{*}{\texttt{Llama-3.2-3B}} & Baseline (src-prompt) & $\underline{.028}_{\pm.125}$ & $.213_{\pm.143}$ & $.273_{\pm.065}$ & $.204_{\pm.103}$ & $\underline{.144}_{\pm.093}$ \\
 & + vector steer (src-prompt) & \textcolor{red!70!black}{$.026_{\pm.127}$$\downarrow$} & \textcolor{blue!70!black}{$\underline{.215}_{\pm.141}$$\uparrow$} & \textcolor{blue!70!black}{$\mathbf{\underline{.278}}_{\pm.064}$$\uparrow$} & \textcolor{blue!70!black}{$\mathbf{\underline{.212}}_{\pm.100}$$\uparrow$} & \textcolor{red!70!black}{$.143_{\pm.090}$$\downarrow$} \\
\midrule
\multirow{2}{*}{\texttt{aya-expanse-8b}} & Baseline (src-prompt) & $.004_{\pm.145}$ & $.129_{\pm.079}$ & $.195_{\pm.065}$ & $.154_{\pm.069}$ & $.158_{\pm.068}$ \\
 & + vector steer (src-prompt) & \textcolor{blue!70!black}{$\underline{.006}_{\pm.144}$$\uparrow$} & \textcolor{blue!70!black}{$\underline{.137}_{\pm.081}$$\uparrow$} & \textcolor{blue!70!black}{$\underline{.200}_{\pm.065}$$\uparrow$} & \textcolor{blue!70!black}{$\underline{.160}_{\pm.070}$$\uparrow$} & \textcolor{blue!70!black}{$\underline{.161}_{\pm.067}$$\uparrow$} \\
\midrule
\multicolumn{7}{l}{\textbf{Direct Prompting}} \\
\midrule
\multirow{2}{*}{\texttt{Llama-3.2-3B}} & Baseline (src-prompt) & $.046_{\pm.039}$ & $\underline{.029}_{\pm.038}$ & $\underline{.056}_{\pm.035}$ & $.081_{\pm.093}$ & $\underline{.115}_{\pm.050}$ \\
 & + vector steer (src-prompt) & \textcolor{blue!70!black}{$\underline{.051}_{\pm.043}$$\uparrow$} & \textcolor{red!70!black}{$.028_{\pm.025}$$\downarrow$} & \textcolor{blue!70!black}{$\underline{.056}_{\pm.034}$$\uparrow$} & \textcolor{blue!70!black}{$\underline{.083}_{\pm.094}$$\uparrow$} & \textcolor{red!70!black}{$.114_{\pm.053}$$\downarrow$} \\
\midrule
\multirow{2}{*}{\texttt{aya-expanse-8b}} & Baseline (src-prompt) & $\mathbf{\underline{.215}}_{\pm.083}$ & $\mathbf{\underline{.244}}_{\pm.148}$ & $\underline{.218}_{\pm.108}$ & $\underline{.188}_{\pm.080}$ & $.215_{\pm.056}$ \\
 & + vector steer (src-prompt) & \textcolor{red!70!black}{$.214_{\pm.085}$$\downarrow$} & \textcolor{blue!70!black}{$\mathbf{\underline{.244}}_{\pm.148}$$\uparrow$} & \textcolor{red!70!black}{$\underline{.218}_{\pm.108}$$\downarrow$} & \textcolor{blue!70!black}{$\underline{.189}_{\pm.079}$$\uparrow$} & \textcolor{blue!70!black}{$\mathbf{\underline{.216}}_{\pm.056}$$\uparrow$} \\
\bottomrule
\end{tabular}%
}
\caption{SEAHORSE Pearson $r$ averaged over 6 quality dimensions (Comprehensible, Non-repetitive, Grammatical, Attributable, Main ideas, Concise) with standard deviation as subscripts ($\text{mean}_{\pm\text{std}}$), per language, using source-language prompts. Baseline uses $\sigma{=}0$ (no steering); steered uses the best $\sigma$ selected on the validation split. \textbf{Bold} indicates the single highest value per language across the entire table; \underline{underline} indicates the highest value per language within each model. \textcolor{blue!70!black}{Blue}/$\uparrow$ (\textcolor{red!70!black}{red}/$\downarrow$) on steered rows indicates improvement (degradation) relative to the baseline.}
\label{tab:seahorse_condensed_src}
\end{table*}
\clearpage

\section{AUC-ROC Results}
\label{app:auc_results}

Tables~\ref{tab:seahorse_condensed_en_auc} and~\ref{tab:seahorse_condensed_src_auc} present AUC-ROC results for the small models
under English and source-language prompting respectively, corresponding to the Pearson~$r$ tables in the main paper
(Table~\ref{tab:seahorse_condensed_en}) and Appendix~\ref{app:src_results}.
Table~\ref{tab:seahorse_condensed_avgvec_large_en_auc} shows AUC-ROC for the large-model avgvec zero-shot setting.

\begin{table*}[t]
\centering
\resizebox{\textwidth}{!}{%
\begin{tabular}{ll|c|c|c|c|c}
\textbf{} & \textbf{} & \textbf{de} & \textbf{es} & \textbf{ru} & \textbf{tr} & \textbf{vi} \\
\toprule
\multicolumn{7}{l}{\textbf{GPTScore}} \\
\midrule
\multirow{2}{*}{\texttt{Llama-3.2-3B}} & Baseline (EN-prompt) & $\underline{.483}_{\pm.112}$ & $.631_{\pm.080}$ & $.698_{\pm.024}$ & $\underline{.628}_{\pm.069}$ & $\underline{.604}_{\pm.051}$ \\
 & + vector steer (EN-prompt) & \textcolor{red!70!black}{$.483_{\pm.111}$$\downarrow$} & \textcolor{blue!70!black}{$\underline{.632}_{\pm.082}$$\uparrow$} & \textcolor{blue!70!black}{$\mathbf{\underline{.702}}_{\pm.024}$$\uparrow$} & \textcolor{red!70!black}{$.626_{\pm.069}$$\downarrow$} & \textcolor{red!70!black}{$.603_{\pm.052}$$\downarrow$} \\
\midrule
\multirow{2}{*}{\texttt{aya-expanse-8b}} & Baseline (EN-prompt) & $.457_{\pm.135}$ & $.597_{\pm.046}$ & $.679_{\pm.055}$ & $.625_{\pm.060}$ & $.641_{\pm.063}$ \\
 & + vector steer (EN-prompt) & \textcolor{blue!70!black}{$\underline{.459}_{\pm.136}$$\uparrow$} & \textcolor{blue!70!black}{$\underline{.600}_{\pm.046}$$\uparrow$} & \textcolor{blue!70!black}{$\underline{.684}_{\pm.056}$$\uparrow$} & \textcolor{blue!70!black}{$\underline{.631}_{\pm.059}$$\uparrow$} & \textcolor{blue!70!black}{$\underline{.643}_{\pm.064}$$\uparrow$} \\
\midrule
\multicolumn{7}{l}{\textbf{Direct Prompting}} \\
\midrule
\multirow{2}{*}{\texttt{Llama-3.2-3B}} & Baseline (EN-prompt) & $\underline{.551}_{\pm.044}$ & $\underline{.497}_{\pm.027}$ & $.510_{\pm.034}$ & $\underline{.499}_{\pm.026}$ & $.502_{\pm.035}$ \\
 & + vector steer (EN-prompt) & \textcolor{red!70!black}{$.550_{\pm.043}$$\downarrow$} & \textcolor{blue!70!black}{$\underline{.498}_{\pm.027}$$\uparrow$} & \textcolor{blue!70!black}{$\underline{.511}_{\pm.033}$$\uparrow$} & \textcolor{blue!70!black}{$\underline{.499}_{\pm.025}$$\uparrow$} & \textcolor{blue!70!black}{$\underline{.505}_{\pm.034}$$\uparrow$} \\
\midrule
\multirow{2}{*}{\texttt{aya-expanse-8b}} & Baseline (EN-prompt) & $.650_{\pm.064}$ & $\mathbf{\underline{.669}}_{\pm.098}$ & $\underline{.676}_{\pm.067}$ & $\mathbf{\underline{.662}}_{\pm.085}$ & $.676_{\pm.055}$ \\
 & + vector steer (EN-prompt) & \textcolor{blue!70!black}{$\mathbf{\underline{.651}}_{\pm.063}$$\uparrow$} & \textcolor{red!70!black}{$\mathbf{\underline{.668}}_{\pm.097}$$\downarrow$} & \textcolor{red!70!black}{$.674_{\pm.066}$$\downarrow$} & \textcolor{red!70!black}{$.661_{\pm.086}$$\downarrow$} & \textcolor{blue!70!black}{$\mathbf{\underline{.678}}_{\pm.055}$$\uparrow$} \\
\bottomrule
\end{tabular}%
}
\caption{SEAHORSE AUC-ROC averaged over 6 quality dimensions (Comprehensible, Non-repetitive, Grammatical, Attributable, Main ideas, Concise) with standard deviation as subscripts ($\text{mean}_{\pm\text{std}}$), per language, using English prompts. Baseline uses $\sigma{=}0$ (no steering); steered uses the best $\sigma$ selected on the validation split. \textbf{Bold} indicates the single highest value per language across the entire table; \underline{underline} indicates the highest value per language within each model. \textcolor{blue!70!black}{Blue}/$\uparrow$ (\textcolor{red!70!black}{red}/$\downarrow$) on steered rows indicates improvement (degradation) relative to the baseline.}
\label{tab:seahorse_condensed_en_auc}
\end{table*}

\begin{table*}[t]
\centering
\resizebox{\textwidth}{!}{%
\begin{tabular}{ll|c|c|c|c|c}
\textbf{} & \textbf{} & \textbf{de} & \textbf{es} & \textbf{ru} & \textbf{tr} & \textbf{vi} \\
\toprule
\multicolumn{7}{l}{\textbf{GPTScore}} \\
\midrule
\multirow{2}{*}{\texttt{Llama-3.2-3B}} & Baseline (src-prompt) & $\underline{.484}_{\pm.108}$ & $.631_{\pm.083}$ & $.697_{\pm.026}$ & $\mathbf{\underline{.631}}_{\pm.065}$ & $\underline{.607}_{\pm.052}$ \\
 & + vector steer (src-prompt) & \textcolor{red!70!black}{$.482_{\pm.112}$$\downarrow$} & \textcolor{blue!70!black}{$\underline{.633}_{\pm.082}$$\uparrow$} & \textcolor{blue!70!black}{$\mathbf{\underline{.701}}_{\pm.027}$$\uparrow$} & \textcolor{red!70!black}{$\mathbf{\underline{.631}}_{\pm.066}$$\downarrow$} & \textcolor{red!70!black}{$.607_{\pm.050}$$\downarrow$} \\
\midrule
\multirow{2}{*}{\texttt{aya-expanse-8b}} & Baseline (src-prompt) & $.451_{\pm.138}$ & $.582_{\pm.041}$ & $.651_{\pm.059}$ & $.600_{\pm.060}$ & $.627_{\pm.061}$ \\
 & + vector steer (src-prompt) & \textcolor{blue!70!black}{$\underline{.453}_{\pm.137}$$\uparrow$} & \textcolor{blue!70!black}{$\underline{.588}_{\pm.043}$$\uparrow$} & \textcolor{blue!70!black}{$\underline{.655}_{\pm.060}$$\uparrow$} & \textcolor{blue!70!black}{$\underline{.604}_{\pm.059}$$\uparrow$} & \textcolor{blue!70!black}{$\underline{.629}_{\pm.062}$$\uparrow$} \\
\midrule
\multicolumn{7}{l}{\textbf{Direct Prompting}} \\
\midrule
\multirow{2}{*}{\texttt{Llama-3.2-3B}} & Baseline (src-prompt) & $.533_{\pm.031}$ & $.529_{\pm.032}$ & $\underline{.550}_{\pm.030}$ & $.551_{\pm.069}$ & $\underline{.575}_{\pm.041}$ \\
 & + vector steer (src-prompt) & \textcolor{blue!70!black}{$\underline{.539}_{\pm.037}$$\uparrow$} & \textcolor{blue!70!black}{$\underline{.529}_{\pm.031}$$\uparrow$} & \textcolor{blue!70!black}{$\underline{.551}_{\pm.029}$$\uparrow$} & \textcolor{blue!70!black}{$\underline{.552}_{\pm.068}$$\uparrow$} & \textcolor{red!70!black}{$.575_{\pm.043}$$\downarrow$} \\
\midrule
\multirow{2}{*}{\texttt{aya-expanse-8b}} & Baseline (src-prompt) & $\mathbf{\underline{.621}}_{\pm.030}$ & $\mathbf{\underline{.656}}_{\pm.102}$ & $\underline{.631}_{\pm.057}$ & $\underline{.611}_{\pm.047}$ & $.631_{\pm.040}$ \\
 & + vector steer (src-prompt) & \textcolor{blue!70!black}{$\mathbf{\underline{.621}}_{\pm.029}$$\uparrow$} & \textcolor{red!70!black}{$\mathbf{\underline{.656}}_{\pm.103}$$\downarrow$} & \textcolor{red!70!black}{$.630_{\pm.057}$$\downarrow$} & \textcolor{blue!70!black}{$\underline{.611}_{\pm.047}$$\uparrow$} & \textcolor{blue!70!black}{$\mathbf{\underline{.632}}_{\pm.038}$$\uparrow$} \\
\bottomrule
\end{tabular}%
}
\caption{SEAHORSE AUC-ROC averaged over 6 quality dimensions (Comprehensible, Non-repetitive, Grammatical, Attributable, Main ideas, Concise) with standard deviation as subscripts ($\text{mean}_{\pm\text{std}}$), per language, using source-language prompts. Baseline uses $\sigma{=}0$ (no steering); steered uses the best $\sigma$ selected on the validation split. \textbf{Bold} indicates the single highest value per language across the entire table; \underline{underline} indicates the highest value per language within each model. \textcolor{blue!70!black}{Blue}/$\uparrow$ (\textcolor{red!70!black}{red}/$\downarrow$) on steered rows indicates improvement (degradation) relative to the baseline.}
\label{tab:seahorse_condensed_src_auc}
\end{table*}

\begin{table*}[t]
\centering
\resizebox{\textwidth}{!}{%
\begin{tabular}{ll|c|c|c|c|c}
\textbf{} & \textbf{} & \textbf{de} & \textbf{es} & \textbf{ru} & \textbf{tr} & \textbf{vi} \\
\toprule
\multicolumn{7}{l}{\textbf{GPTScore}} \\
\midrule
\multirow{2}{*}{\texttt{Llama-3.1-8B-Instruct}} & Baseline (EN-prompt) & $.480_{\pm.120}$ & $.643_{\pm.077}$ & $.711_{\pm.028}$ & $.659_{\pm.059}$ & $\underline{.631}_{\pm.051}$ \\
 & + avgvec zero-shot (EN-prompt) & \textcolor{blue!70!black}{$\mathbf{\underline{.487}}_{\pm.113}$$\uparrow$} & \textcolor{blue!70!black}{$\mathbf{\underline{.649}}_{\pm.084}$$\uparrow$} & \textcolor{blue!70!black}{$\mathbf{\underline{.717}}_{\pm.029}$$\uparrow$} & \textcolor{blue!70!black}{$\mathbf{\underline{.660}}_{\pm.058}$$\uparrow$} & \textcolor{red!70!black}{$.629_{\pm.051}$$\downarrow$} \\
\midrule
\multirow{2}{*}{\texttt{aya-expanse-32b}} & Baseline (EN-prompt) & $.448_{\pm.131}$ & $.603_{\pm.039}$ & $.680_{\pm.048}$ & $.626_{\pm.061}$ & $.642_{\pm.055}$ \\
 & + avgvec zero-shot (EN-prompt) & \textcolor{blue!70!black}{$\underline{.451}_{\pm.131}$$\uparrow$} & \textcolor{blue!70!black}{$\underline{.607}_{\pm.040}$$\uparrow$} & \textcolor{blue!70!black}{$\underline{.683}_{\pm.049}$$\uparrow$} & \textcolor{blue!70!black}{$\underline{.631}_{\pm.060}$$\uparrow$} & \textcolor{blue!70!black}{$\mathbf{\underline{.645}}_{\pm.056}$$\uparrow$} \\
\bottomrule
\end{tabular}%
}
\caption{SEAHORSE AUC-ROC averaged over 6 quality dimensions (Comprehensible, Non-repetitive, Grammatical, Attributable, Main ideas, Concise) with standard deviation as subscripts ($\text{mean}_{\pm\text{std}}$), per language, using English prompts. Baseline uses $\sigma{=}0$ (no steering); avgvec zero-shot uses a fixed steering vector averaged across all source languages with a pre-set $\sigma$. \textbf{Bold} indicates the single highest value per language across the entire table; \underline{underline} indicates the highest value per language within each model. \textcolor{blue!70!black}{Blue}/$\uparrow$ (\textcolor{red!70!black}{red}/$\downarrow$) on steered rows indicates improvement (degradation) relative to the baseline.}
\label{tab:seahorse_condensed_avgvec_large_en_auc}
\end{table*}

\clearpage

\end{document}